\documentclass{article}

\PassOptionsToPackage{numbers, compress}{natbib}

\usepackage[preprint]{neurips_2026}
\usepackage[table]{xcolor}
\usepackage[utf8]{inputenc} 
\usepackage[T1]{fontenc}    
\usepackage{hyperref}       
\usepackage{url}            
\usepackage{booktabs}       
\usepackage{amsfonts}       
\usepackage{nicefrac}       
\usepackage{microtype}      
\usepackage{changes}
\usepackage{graphicx}
\usepackage{subcaption}
\usepackage{animate}
\usepackage{amsmath}

\usepackage{wrapfig}
\usepackage{float}
\usepackage[ruled,vlined]{algorithm2e}
\SetKw{KwBreak}{break}
\SetKw{KwContinue}{continue}
\providecommand{\Clamp}{\mathop{\mathrm{Clamp}}}
\usepackage{amssymb}

\title{Geometric 4D Stitching for Grounded  4D Generation}

\author{
    Sunwoo Park \qquad Taesung Kwon\thanks{Corresponding authors.} \qquad Jong Chul Ye\footnotemark[2] \\
    KAIST AI \\
    \texttt{\{sunwoo\_p, star.kwon, jong.ye\}@kaist.ac.kr}
}

\begin{document}

\maketitle

\begin{figure}[H]
    \centering
    \includegraphics[width=0.9\linewidth]{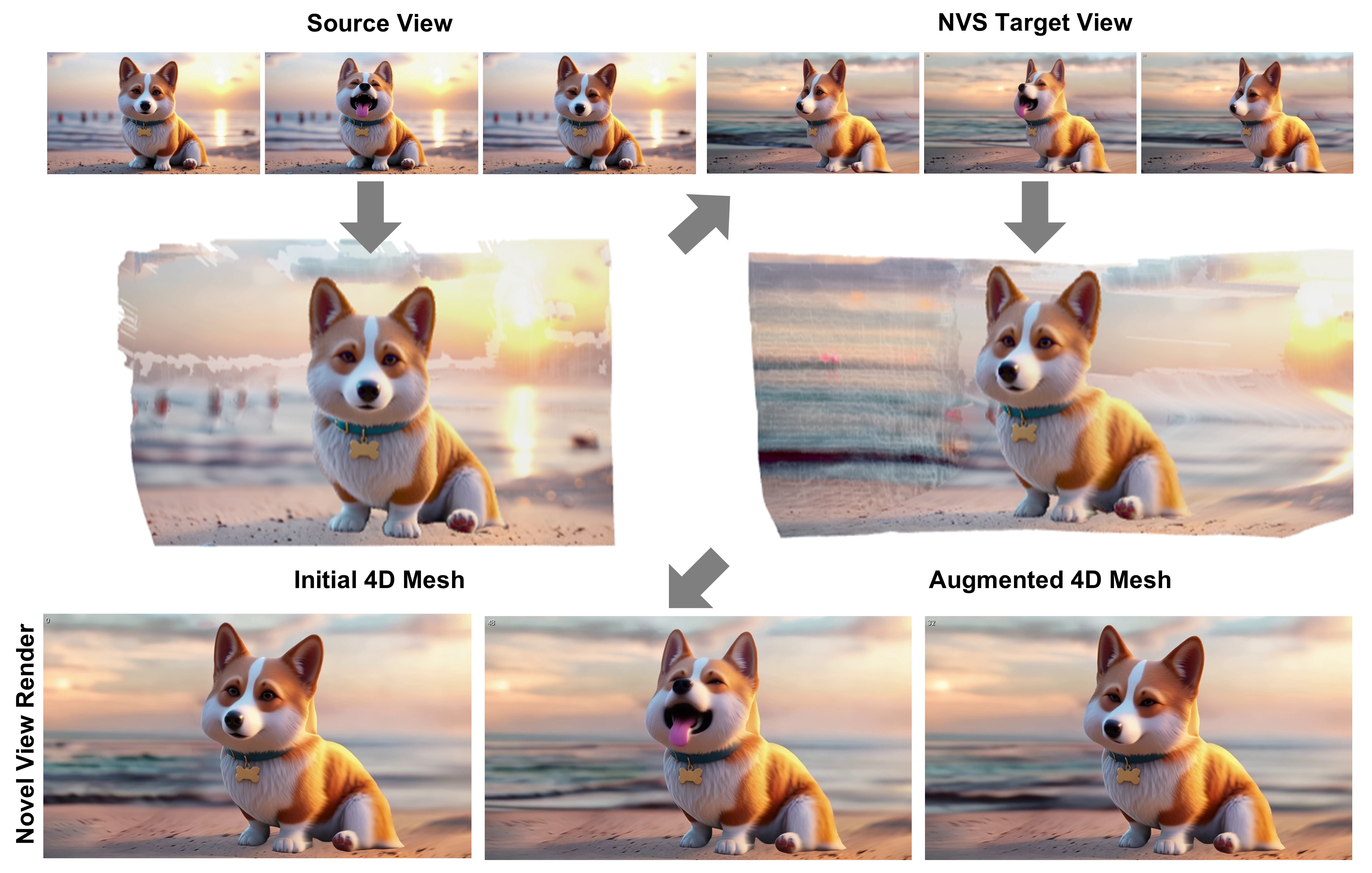}
    \caption{
    We present \textbf{Geometric 4D Stitching}, a geometry-aware approach for constructing expandable 4D scene representations from sparse generative videos.
    Our method refines NVS-generated target-view geometry using the source-view geometry as an anchor, stitches only reliable newly revealed regions, and enables coherent novel view rendering from the augmented 4D scene.
    }
    \label{fig:fig0}
\end{figure}

\label{sec:intro}

\begin{abstract}
Recent 4D generation methods complete scene-level missing information using generative models and reconstruct the scene into radiance-based representations. However, these pipelines often present geometric inconsistencies in the generated content, and the radiance-based reconstruction requires expensive optimization. Furthermore, radiance-based representations often absorb these geometric inconsistencies into their view-dependent nature, failing to enforce the grounded geometric consistency. To address these issues, we propose \textbf{Geometric 4D Stitching}, an efficient framework that explicitly identifies missing geometric regions and complements them with geometrically grounded 4D stitches. As a result, our method constructs 4D scene representations in under 10 minutes on a single NVIDIA RTX 5090 GPU per one-step scene expansion, while improving geometric consistency.
Moreover, we demonstrate that our explicit 4D stitching supports interative expansion of 4D mesh as well as 4D scene editing.
\end{abstract}

\section{Introduction}

Recent 4D generation methods~\cite{miao2025advances4dgeneration,Wu_2025_CVPR,Liu_2025_ICCV,xie2025sv4d} complete scene-level missing information by synthesizing multi-view videos with generative models, and reconstruct the completed information into 4D representations, most commonly through radiance-field optimization, such as Gaussian splatting~\cite{kerbl3Dgaussians, Wu_2024_CVPR} and NeRF~\cite{mildenhall2020nerf,Pumarola_2021_CVPR}.

However, synthesizing multi-view videos with generative models imposes substantial computational overhead in ensuring multi-view consistency and alignment~\cite{park2025zero4d,Wu_2025_CVPR,xie2025sv4d}.
Moreover, even with costly alignment or optimization of current novel-view synthesis (NVS) models~\cite{yu2025trajectorycrafter,bai2025recammaster}, often present geometric inconsistencies in the generated content.
As a result, multi-view videos with geometric inconsistencies can amplify conflicting geometric evidence, making 4D reconstruction an \textit{ill-posed setting}.
In this regime, radiance-based 4D reconstruction may absorb these geometric inconsistencies into their view-dependent nature, rather than recovering a geometrically aligned scene.

To address this issue, we propose \textbf{Geometric 4D Stitching}, an efficient framework that replaces dense generative completion and radiance-based reconstruction with explicit geometric augmentation.
Instead of completing all unseen scene-level information via dense views, our method explicitly identifies incomplete geometric regions and directly generates geometrically aligned 4D stitches to infill them.
This formulation reframes 4D generation, shifting from scene-level generative completion to region-level geometric completion.
By operating only on the regions that require additional information, our method avoids redundant view generation and minimizes the alignment cost.

As a result, our method constructs grounded 4D scene representations that is not tied to a specific camera view and enables consistent novel-view rendering, in under 10 minutes on a single NVIDIA RTX 5090 GPU.
Furthermore, we demonstrate that this explicit 4D representation enables practical downstream applications that are difficult to achieve with radiance-based pipelines, including geometrically aligned scene extrapolation and 4D scene editing.

We summarize our contributions as follows:
\begin{itemize}
    \item \textbf{Identifying a limitation of dense generative completion for 4D generation.}
    We point out that increasing observation density for reconstruction can increase overlap among NVS-generated views, which may rapidly accumulate multi-view consistency and alignment costs under the limited capacity of current NVS models.
    \item \textbf{Geometric 4D Stitching.}
    We reformulate 4D generation as an explicit geometric stitching problem. Instead of dense scene-level completion, our method enables region-level geometric completion by integrating geometrically aligned 4D \textbf{stitches} into \textbf{incomplete regions} of the initial 4D asset.
    \item \textbf{Practical Downstream Applications.}
    We demonstrate that our explicit 4D representation enables practical tasks that remain challenging for radiance-based methods, such as geometrically aligned scene extrapolation and 4D scene editing.
\end{itemize}

\section{Related Work}
\label{sec:related}

\paragraph{Novel-View Synthesis.}
Novel-view synthesis (NVS)~\cite{yu2025trajectorycrafter,bai2025recammaster}, is a key primitive for video-based 4D generation.
Most recent NVS methods build on video diffusion models (VDMs)~\cite{yang2025cogvideox,wan2025wan} to ensure consistent video generation, integrating geometric conditions, such as camera trajectories, depth maps, or warped source features, to synthesize an input video from novel viewpoints. 
Building upon this, \textbf{multi-view} video synthesis~\cite{park2025zero4d,xie2025sv4d,Wu_2025_CVPR,Liu_2025_ICCV}, is designed to combine the camera-axis expansion of NVS with VDM priors, synthesizing novel viewpoints while maintaining temporally plausible motion.
Consequently, the resulting multi-view consistency is fundamentally bounded by the capacity of the underlying VDM prior, which is primarily trained to generate visually plausible videos rather than preserving 3D structural consistency.
Recent geometry-aware studies~\cite{sun2026vggtworld,du2026videogpa} further support this gap: visually plausible video predictions can remain geometrically inconsistent, and video diffusion models often struggle to maintain rigid 3D structures under camera motion.
This distinction is critical for 4D generation, where strict geometric consistency is a fundamental prerequisite, rather than just visually plausible outputs.
Our work pinpoints this issue, focusing on enforcing structural consistency by using NVS outputs exclusively to generate geometrically grounded 4D stitches.

\begin{figure}[!t]
    \centering
    \includegraphics[width=1\linewidth]{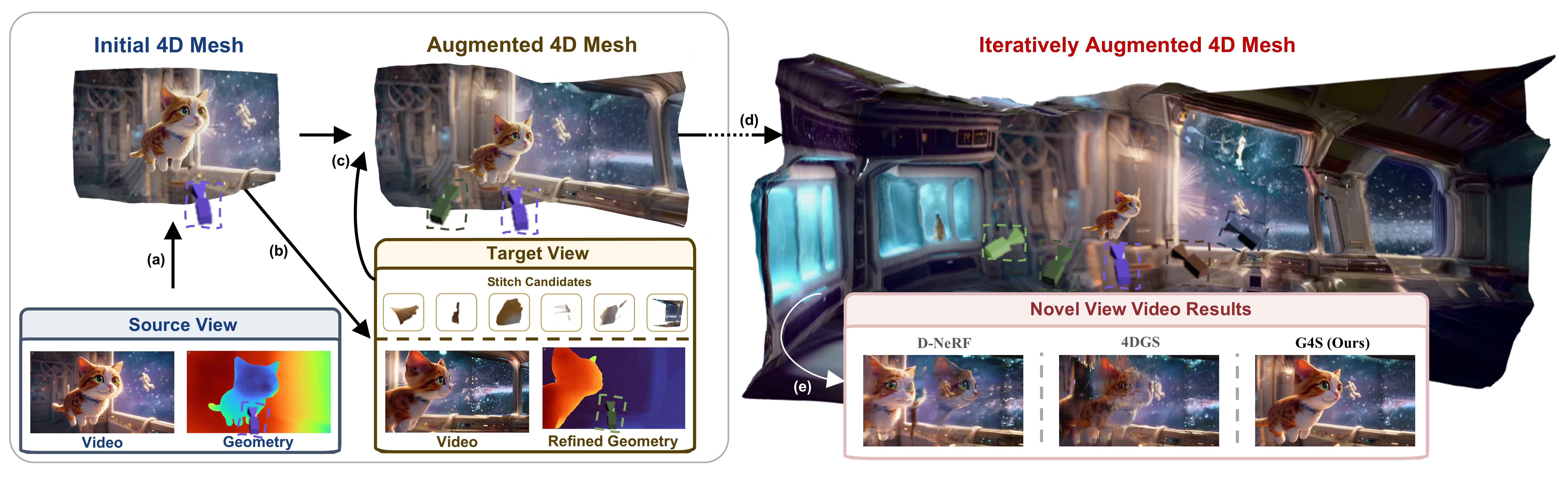}
    \caption{
\textbf{Overview of Geometric 4D Stitching.}
Given a condition video, (a) we first construct an initial 4D mesh and (b) use its geometry as an anchor to refine the geometry of the completed target video.
(c) The resulting information-addition regions are converted into stitch candidates and inserted into the mesh, (d) producing an augmented 4D representation that can be progressively expanded.
Finally, (e) we apply a lightweight NVS-based visual refinement to improve novel-view rendering while preserving the underlying stitched geometry.
Compared with NeRF~\cite{Pumarola_2021_CVPR} and Gaussian Splatting~\cite{Wu_2024_CVPR}, G4S renders more coherent videos from sparse generative observations.
}
    \label{fig:main_figure}
\end{figure}

\paragraph{4D Reconstruction.}
Radiance-based approaches~\cite{mildenhall2020nerf,Pumarola_2021_CVPR,Wu_2024_CVPR} optimize multi-view observations using photometric losses to construct a 4D representation.
However, they often absorb geometric inconsistencies into their view-dependent effects, failing to enforce strict geometric consistency and thereby limiting downstream manipulability~\cite{yuan2022nerfediting,lazova2023controlnerf}.

\begin{wrapfigure}{r}{0.4\textwidth}
  \centering
  \vspace{-8pt}
\includegraphics[width=0.4\textwidth]{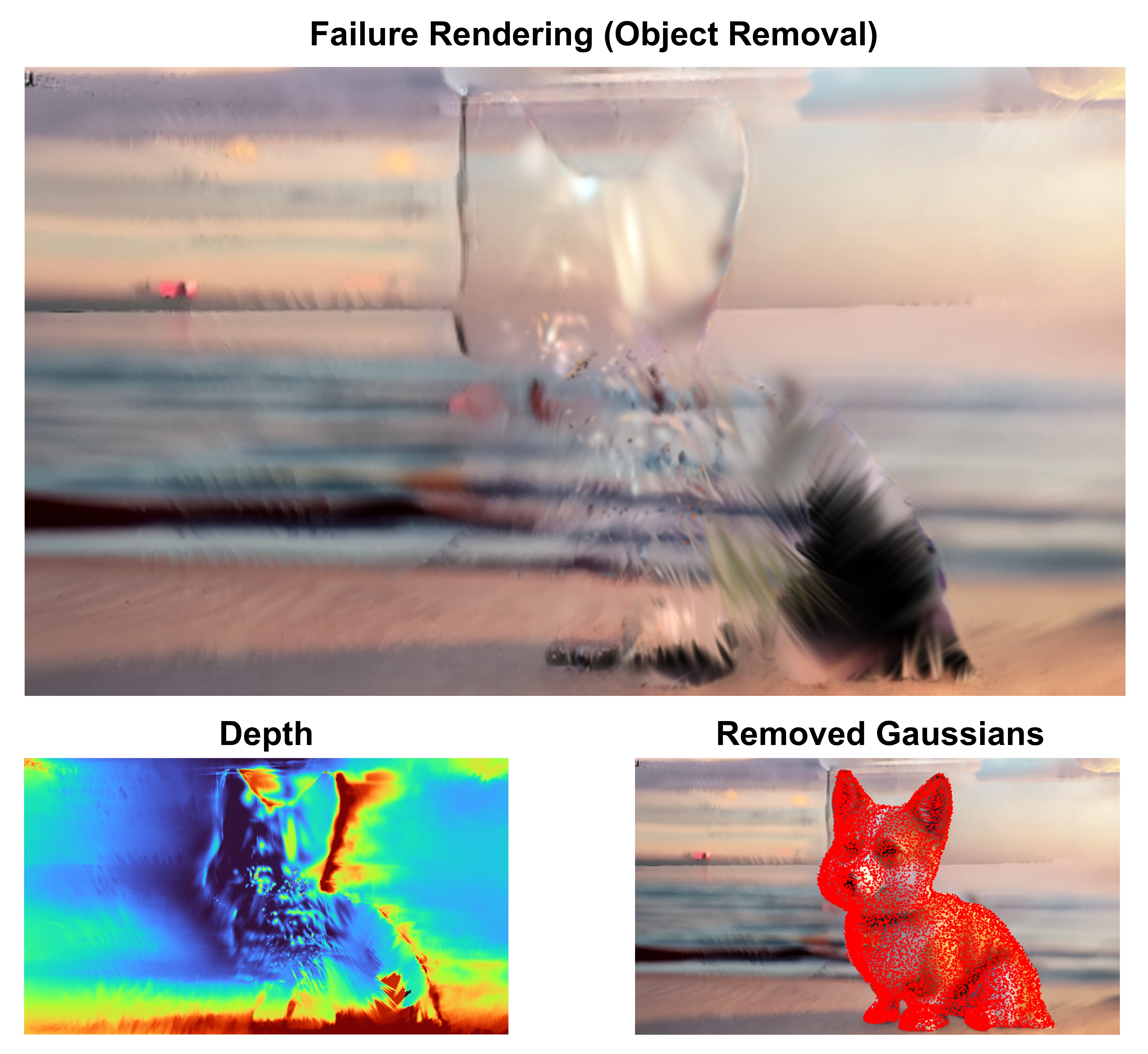}
  \caption{
    \textbf{Failure Case of Gaussian-based Scene Manipulation} due to difficult direct Gaussian matching to object boundaries.}
    \label{fig:fig_fail}
    \vspace{-12pt}
\end{wrapfigure}

To bypass this expensive optimization, recent Feed-Forward Geometry Transformers (FF-GTs), such as VGGT and DA3~\cite{wang2024vggt, yang2025depthanything3}, offer rapid reconstruction.
While this line of work has led to fast 3D/4D reconstruction methods like AnySplat and Instant4D~\cite{jiang2025anysplat, luo2025instant4d}, applying them to generative multi-view settings remains challenging.
Although NVS-generated multi-view videos can be visually plausible, their geometry is often misaligned across views.
Such geometric misalignment leads to camera-depth calibration inconsistencies that severely degrade downstream 4D asset quality.

To mitigate the geometric misalignment issues of previous methods, we utilize the geometry of the initial-view asset as an anchor to refine the generated novel-view content, directly stitching only the required regions.
By geometrically grounding the stitches prior to integration, our method enables highly efficient, explicit, and controllable 4D asset generation.

\section{Method}
\label{sec:method}

In this section, we detail the architecture of our efficient and geometrically consistent 4D generation framework, Geometric 4D Stitching.
We begin by \textbf{generating an initial-view 4D asset} from the input video (Section~\ref{sec:initial_view_4d_asset}).
Given this initial asset, our goal is to expand it using newly revealed geometry from Novel-View Synthesis (NVS) while avoiding redundant view generation.
To achieve this, Section~\ref{sec:information_addition} defines \textbf{information addition regions} via NVS, identifying exactly where new content should be inserted.
Next, in Section~\ref{sec:pyramidal_refinement}, we refine this generated novel-view geometry using \textbf{pyramidal refinement}, carefully aligning it with the initial-view 4D asset.
Finally, Section~\ref{sec:geometric_stitching} introduces \textbf{geometric 4D stitching}, which integrates only reliable, geometry-refined stitch candidates into the asset.
This strategy preserves already reconstructed areas and significantly boosts computational efficiency.

\subsection{Generate the Initial-View 4D Asset}
\label{sec:initial_view_4d_asset}

We first generate an initial-view 4D asset from the source video, which can be either observed or generated.
For each frame $t$, we estimate the source-view camera and depth using DA3~\cite{yang2025depthanything3}.
We then construct a triangular mesh on the image lattice by connecting neighboring valid pixels into faces.
Each mesh vertex is lifted into 3D using the estimated source-view camera and depth, while the corresponding RGB values are assigned as vertex colors or textures.
Repeating this process over time produces a time-indexed raw mesh sequence, which serves as the basic form of the initial explicit 4D asset.

This initial-view 4D asset $\mathcal{G}_{i}^{t}$ provides the geometric reference and visual context required for subsequent NVS.
Through further refinement and stitching, it becomes the foundation for progressive 4D asset expansion. In practice, we also perform background completion at this stage to alleviate occlusion-induced information deficiency under sparse-view observations.

\subsection{Novel View Synthesis with Information-Addition Region}
\label{sec:information_addition}

\begin{figure}
    \centering
    \includegraphics[width=1\linewidth]{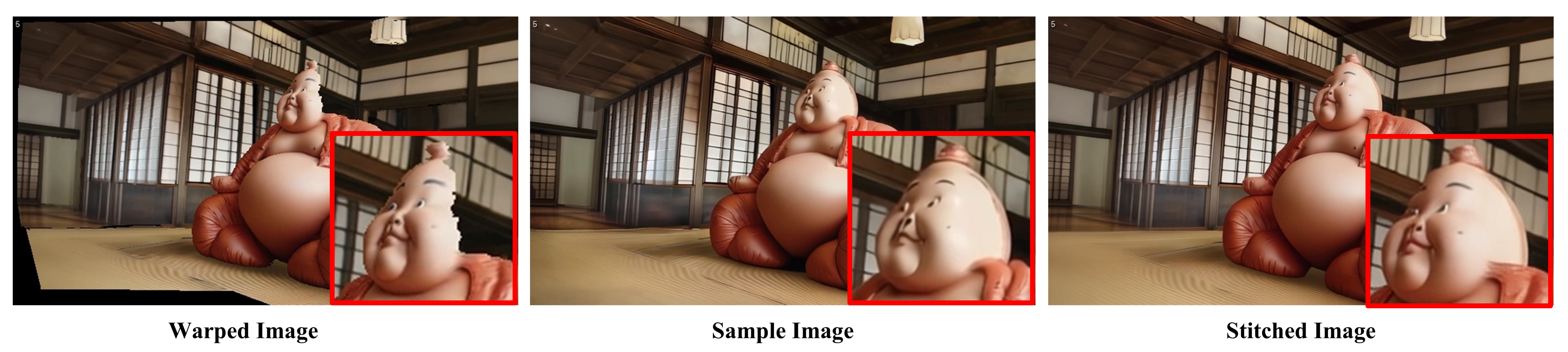}
\caption{
\textbf{Stitching of warped and sampled images.}
The warped image preserves source-supported structure but contains invalid regions, while the sampled image fills these areas at the cost of altering some source content. The stitched image retains the reliable projected structure and incorporates generated content only where needed, yielding a more coherent final result.
}
    \label{fig:fig2_r}
\end{figure}

From initial-view 4D asset, we generate target-view to identify information-addition region.
Given a source view $i$ at time $t$, we render its RGB and depth into a target view $j$ using pointcloud rendering, obtaining a projected image $I^{t}_{i\rightarrow j}$ and a projection-valid mask $\Omega^{\mathrm{t}}_{i\rightarrow j}$.
We treat only the projection-invalid pixels as missing regions and fills them with the NVS module $\Phi_{\mathrm{nvs}}$:
\begin{equation}
\hat{I}^{t}_{i\rightarrow j}
=
\Phi_{\mathrm{nvs}}
\left(
I^{t}_{i\rightarrow j},
M^{\mathrm{t}}_{i\rightarrow j}
\right)
,
\qquad
M^{\mathrm{t}}_{i\rightarrow j}
=
1-\Omega^{\mathrm{t}}_{i\rightarrow j}.
\end{equation}

\begin{wrapfigure}{r}{0.35\textwidth}
  \centering
\includegraphics[width=0.35\textwidth]{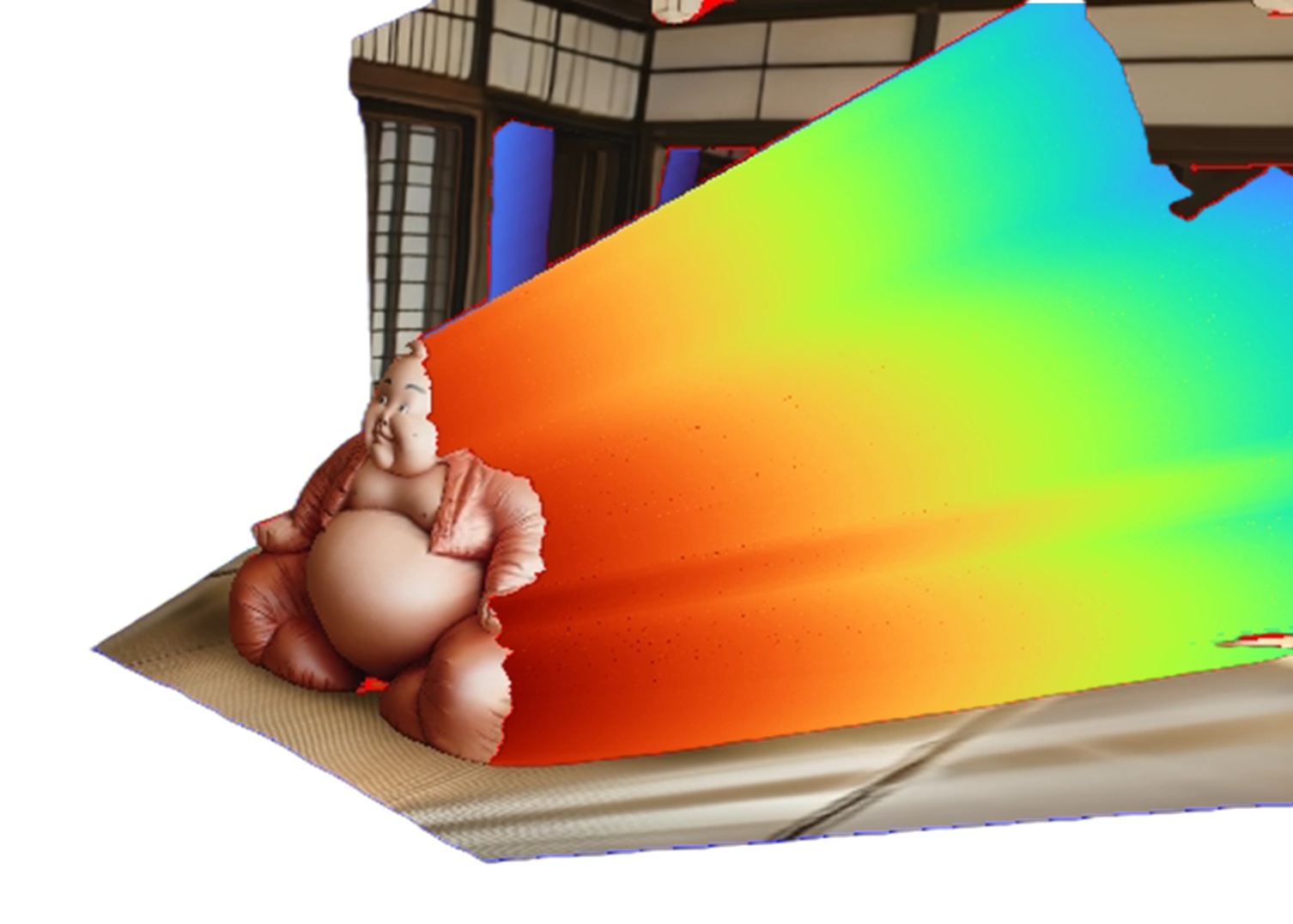}
  \caption{
    \textbf{Curtain and its depth.}
    We identify information-addition regions from the depth discrepancy between raw mesh and point-cloud renderings.}
    \label{fig:fig_curtain}
\end{wrapfigure}

Here, $M^{\mathrm{t}}_{i\rightarrow j}$ denotes the inpainting mask based on $\Omega^{\mathrm{t}}_{i\rightarrow j}$, and $\Phi_{\mathrm{nvs}}$ takes a projected image and an inpainting mask and outputs a full novel-views.

However, projection holes alone are not sufficient.
Regions occluded in the source view can become newly visible in the target view, yet under standard projection they may appear partially supported by other projected source pixels.
To explicitly identify these geometry-missing regions, we compare renderings of the raw 4D mesh and its point cloud.
While raw meshes often incorrectly connect neighboring points across occlusion boundaries, forming "curtains" (Fig.~\ref{fig:fig_curtain}), point clouds do not suffer from this incorrect connection.
This discrepancy directly exposes occlusion-induced missing regions, denoted as $M^{t,+}_{i\rightarrow j}$.
By adding this to the original inpainting mask, we define our enlarged information-addition mask as
$
\tilde{M}^{\mathrm{t}}_{i\rightarrow j}
=
M^{\mathrm{t}}_{i\rightarrow j}
\cup
M^{\mathrm{t},+}_{i\rightarrow j}.
$
We then apply NVS explicitly over this region to extract candidate content:
\begin{equation}
\hat{I}^{t}_{i\rightarrow j}
=
\Phi_{\mathrm{nvs}}
\left(
I^{t}_{i\rightarrow j},
\tilde{M}^{\mathrm{t}}_{i\rightarrow j}
\right),
\qquad
C^{t}_{i\rightarrow j}
=
\tilde{M}^{\mathrm{t}}_{i\rightarrow j}
\odot
\hat{I}^{t}_{i\rightarrow j}.
\end{equation}
Importantly, we do not directly integrate this NVS output into the 4D asset.
Instead, we pass only the candidate pixels $C^{t}_{i\rightarrow j}$ to the final stage, ensuring that only reliable regions are stitched. 
Further details for the NVS module can be found in App.~\ref{app_nvs}.

\subsection{Geometric Alignment with Pyramidal Refinement}
\label{sec:pyramidal_refinement}

\begin{figure}
    \centering \includegraphics[width=0.8\linewidth]{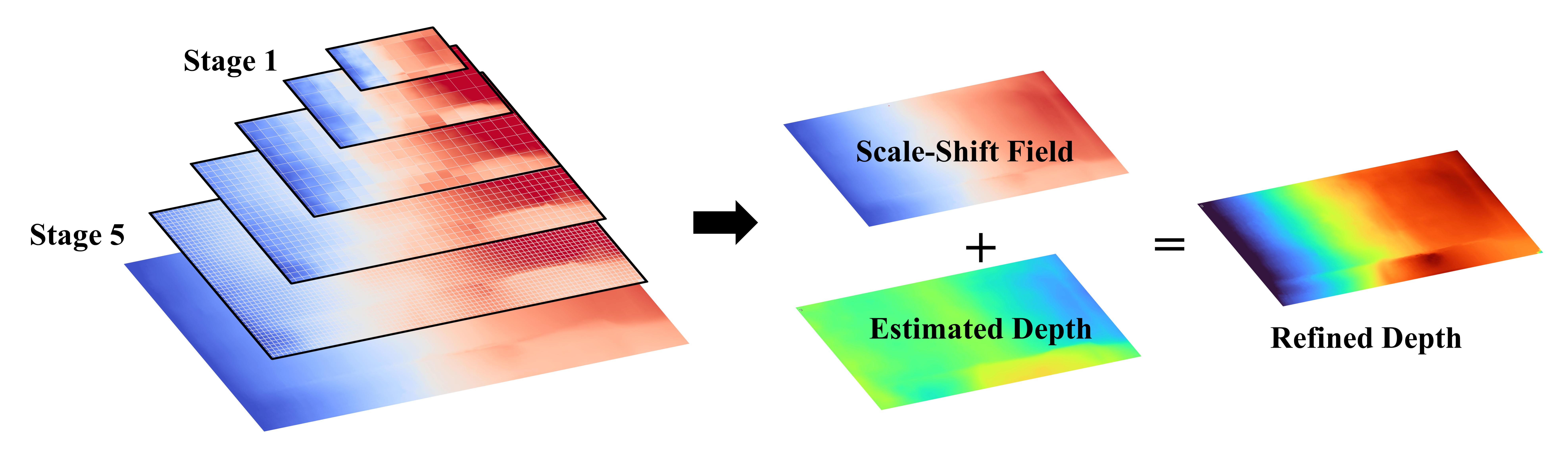}
    \caption{\textbf{Geometric Alignmnet with Pyramidal Refinement.} We construct the continuous scale-shift field constrained to scene geometry for appropriate depth refinement. Details are in App~\ref{depth_detail}.}
    \label{fig:depth_refine2}
\end{figure}

For geometrically grounded 4D stitching, the generated novel-view, or target-view, depth should be aligned with the initial mesh geometry.
However, the mismatch between the generated depth and the initial-view depth varies spatially, making a simple global scale-shift correction insufficient, as shown in Fig.~\ref{fig:depth_refine}.
Furthermore, directly relying on off-the-shelf depth completion models does not provide sufficiently plausible and temporally consistent geometry for 4D asset construction, as shown in Fig.~\ref{fig:fig10}.
We therefore formulate depth refinement as a local scale-shift field $\Theta(\mathbf{x})$ correction:
\begin{equation}
\Theta(\mathbf{x}) = \bigl(s(\mathbf{x}), b(\mathbf{x})\bigr),
\qquad
\tilde{D}(\mathbf{x};\Theta)
=
s(\mathbf{x})\hat{D}(\mathbf{x}) + b(\mathbf{x}),
\end{equation}
where $\hat{D}$ is the generated depth generated and $\tilde{D}$ is the refined depth.
Rather than estimating $\Theta$ for every pixel, we estimate it patch-wise and optimize them with \textbf{pyramidal refinement}~(Fig.\ref{fig:depth_refine2}), first correcting large depth misalignment and then refining local geometric details later.

The refinement objective consists of three terms: anchor alignment, geometry-aware smoothness in all patches $\mathcal{E}$, and stabilization in non-anchor patches $\mathcal{E}_{\mathrm{non-anc}}$:
\begin{equation}
\begin{aligned}
\mathcal{L}(\Theta)
=&
\lambda_1
\sum_{\mathbf{x}\in\mathcal{M}_{\mathrm{anc}}}
\left\|
\tilde{D}(\mathbf{x};\Theta)
-
D_{\mathrm{anc}}(\mathbf{x})
\right\|_1 \\
&+
\lambda_2
\sum_{(i,j)\in\mathcal{E}}
w_{ij}
\left\|
\Theta_i-\Theta_j
\right\|_1
+
\lambda_3
\sum_{(i,j)\in\mathcal{E}_{\mathrm{non\text{-}anc}}}
w_{ij}
\left\|
\Theta_i-\Theta_j
\right\|_1 .
\end{aligned}
\end{equation}
Here, $D_{\mathrm{anc}}$ denotes the anchor depth rendered from the existing asset, and $\mathcal{M}_{\mathrm{anc}}$ is the pixel-level region where the anchor depth is valid.
The affinity $w_{ij}$ is computed from geometry cues such as surface normals or depth boundaries, allowing correction to propagate within coherent surfaces while suppressing propagation across discontinuities or object boundaries.
This refinement aligns the NVS-generated geometry with the initial asset.
By reducing depth discrepancies, it provides more reliable novel-view geometry for the subsequent stitching stage. Further details are in App.~\ref{depth_detail}.

\subsection{Geometric 4D Stitching}
\label{sec:geometric_stitching}

After geometry refinement, we generate stitch from candidate content into an explicit 4D representation.
Specifically, we do not use the full NVS output as supervision.
Instead, we insert only the regions selected by the information-addition mask $\tilde{M}_{i\rightarrow j}$ and supported by the refined novel (or target)-view geometry.

For each source-target pair $(i,j)$ at time $t$, we use the information-addition mask $\tilde{M}_{i\rightarrow j}$ from Sec.~\ref{sec:information_addition} as the base insertion region. We restrict insertion to the information-addition area $\tilde{M}_{i\rightarrow j}$ and,
using the refined target-view depth $\tilde D^t_{i\rightarrow j}$ from Sec.~\ref{sec:pyramidal_refinement}, we back-project the selected target-view pixels to form candidate additions:
\begin{equation}
\mathcal{G}_{i\rightarrow j}^{t}
=
\left\{
\left(
\Pi^{-1}\!\left(\pi_j^{t},\, p,\, \tilde D^t_{i\rightarrow j}(p)\right),
\hat I^t_{i\rightarrow j}(p)
\right)
\;\middle|\;
p\in \tilde{M}_{i\rightarrow j}
\right\}.
\label{eq:stitch_candidates}
\end{equation}
Thus, the NVS output contributes only through geometry-refined candidate regions, rather than being used as dense reconstruction supervision.
We then augment the initial asset geometry $\mathcal{G}_i$ by adding the candidate geometry $\mathcal{G}_{i\rightarrow j}^{t}$, producing the final stitched representation:
\begin{equation}
\mathcal{G}_{\mathrm{final}}^{t}
=
\mathcal{G}_{i}^{t}
\cup
\mathcal{G}_{i\rightarrow j}^{t}.
\end{equation}

\begin{figure}
    \centering
    \includegraphics[width=1\linewidth]{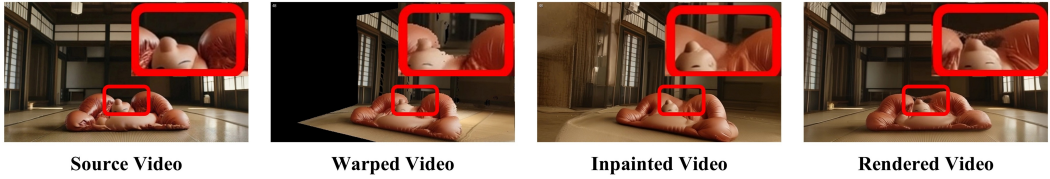}
    \caption{
\textbf{Conflicting evidence from naive stitching of NVS outputs.}
When novel-view synthesis generates structures that are not supported by the source observation, directly stitching them into the existing 4D scene introduces conflicting geometry rather than consistent scene expansion.
}
    \label{fig:fig150}
\end{figure}

\paragraph{\textbf{Render-disagreement distillation.}}
Since NVS does not explicitly guarantee geometric consistency, some candidate additions may still conflict with the existing asset~(Fig.~\ref{fig:fig150}).
To mitigate this, we render the scene after tentative stitching back to the observed views and compare it with an anchor render obtained from the original asset geometry.
If a newly added candidate consistently disagrees with the anchor render across observed views, we regard it as likely to contaminate the existing asset and remove it.
This filtering step preserves valid newly revealed geometry while suppressing unstable additions caused by generated NVS content.

\subsection{Visual Refinement of Explicit Geometry Rendering}
\label{sec:visual_refinement}

After we integrate gemetry-refince stitch candidates into the initial 4D asset, we get augmented 4D mesh $\mathcal{G}_{\mathrm{final}}^{t}$ and now we are ready to render this 4D mesh to any camera trajectories.
For mesh rendering, we apply a lightweight diffusion-based refinement step, to improve visual quality without altering the reconstructed geometry.
Specifically, we use TrajectoryCrafter~\cite{yu2025trajectorycrafter} as a single-step refinement module conditioned on the mesh rendering.

\begin{figure}
    \centering
    \includegraphics[width=1\linewidth]{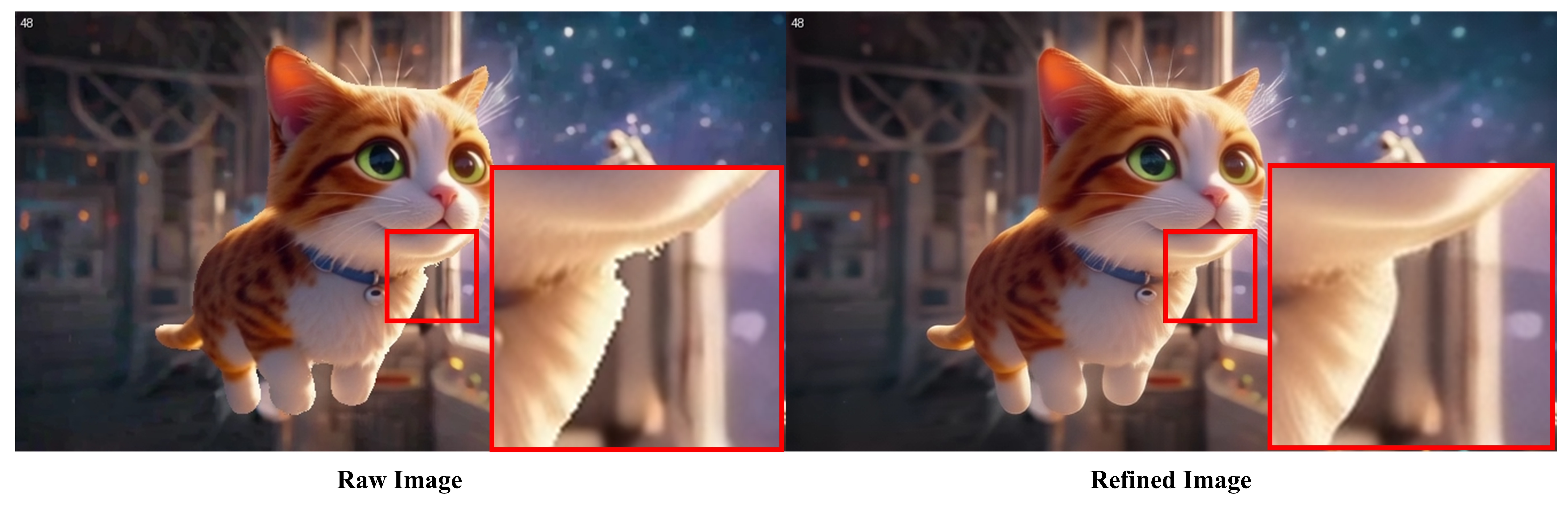}
        \caption{
    \textbf{Geometry-preserving visual refinement.}
    Single-step TrajectoryCrafter refinement improves raw mesh renderings by reducing boundary artifacts, while preserving the underlying geometry.
    }
    \label{fig:fig3_r}
\end{figure}

We pass the raw rendering of our explicit 4D representation as the conditioning image and applying only one sampling step.
This refinement mainly reduces high-frequency artifacts in the rendered video as shown in Fig.~\ref{fig:fig3_r}.
The refined results improve visual fidelity while preserving the underlying geometry of the stitched 4D representation.

\section{Experiments}

Our experiments are designed to evaluate whether the proposed geometry-aware stitching pipeline can construct coherent 4D scene assets from sparse generative observations.
Accordingly, we evaluate our method along three complementary axes.
First, we measure perceptual video quality to verify that the rendered novel views remain visually plausible.
Second, we evaluate camera-motion fidelity as a proxy for geometric robustness by rendering videos from the reconstructed 4D scene along the input camera trajectory and re-estimating their camera poses.
A geometrically consistent 4D scene should yield estimated trajectories that agree with the original input/render trajectory.
Third, we assess geometry--motion self-consistency through reprojection-based evaluation.

\paragraph{Experimental setup.}
We evaluate whether the proposed geometry-aware stitching pipeline can construct coherent 4D scene assets from sparse generative observations.
We generate source scenes with SORA2~\cite{openai_sora2} and use TrajectoryCrafter~\cite{yu2025trajectorycrafter} as the NVS backbone.
Rather than treating the full NVS output as reconstruction supervision, our method uses NVS only to synthesize candidate content in the information-addition regions.
We use DA3~\cite{yang2025depthanything3} as the feed-forward geometry estimator.
The generated candidate content is refined by the proposed geometry refinement stage and then selectively stitched into the explicit 4D representation.
This setup directly tests the central design of our method: whether generated novel-view content can be used safely when it is restricted to necessary regions, geometrically refined, and selectively inserted.
We compare against representative radiance-field optimization baselines, D-NeRF~\cite{Pumarola_2021_CVPR} and 4DGS~\cite{Wu_2024_CVPR}, under the same sparse-view setting.
Implementation details such as resolution, frame settings, mask post-processing, prompting, rendering refinements are provided in the Appendix.

\subsection{Perceptual Video Quality via VBench}
\label{sec:eval_vbench}
We evaluate perceptual quality and temporal plausibility with VBench~\cite{huang2024vbench} on our sparse-view novel-view rendering setup (Tab~\ref{tab:vbench}).  
For each scene, we compute VBench scores on the rendered videos and report per-dimension results as well as the overall average (higher is better). 
These dimensions diagnose complementary failure modes: Subject consistency, Background consistency, Motion smoothness, Aesthetic quality, Image quality.
We additionally report the mean score as the average across dimensions to provide a concise summary of overall perceptual performance.

\begin{table}[tb]
  \caption{\textbf{VBench Result} : Perceptual video quality}
  \label{tab:vbench}
  \centering
  \setlength{\tabcolsep}{3pt}
  \renewcommand{\arraystretch}{1.05}
  \resizebox{\textwidth}{!}{
  \begin{tabular}{@{}ll|lll|lll|lll|lll|lll@{}}
    \toprule
    \multicolumn{2}{c|}{\textbf{Scene}} 
      & \multicolumn{3}{c|}{\textbf{Buddha}}
      & \multicolumn{3}{c|}{\textbf{Cat}}
      & \multicolumn{3}{c|}{\textbf{Corgi}}
      & \multicolumn{3}{c|}{\textbf{Frankenstein}}
      & \multicolumn{3}{c}{\textbf{Mean}} \\
    \midrule
    \multicolumn{1}{c}{\textbf{Metric / Method}} & 
      & D-NeRF & 4DGS & \cellcolor{yellow!25}\textbf{Ours}
      & D-NeRF & 4DGS & \cellcolor{yellow!25}\textbf{Ours}
      & D-NeRF & 4DGS & \cellcolor{yellow!25}\textbf{Ours}
      & D-NeRF & 4DGS & \cellcolor{yellow!25}\textbf{Ours}
      & D-NeRF & 4DGS & \cellcolor{yellow!25}\textbf{Ours} \\
    \midrule

    \multicolumn{2}{l|}{\textbf{Subject Consistency}}
      & 0.797 & \underline{0.798} & \cellcolor{yellow!25}\textbf{0.900}
      & \underline{0.912} & 0.883 & \cellcolor{yellow!25}\textbf{0.959}
      & 0.616 & \underline{0.825} & \cellcolor{yellow!25}\textbf{0.966}
      & 0.859 & \underline{0.869} & \cellcolor{yellow!25}\textbf{0.921}
      & 0.796 & \underline{0.844} & \cellcolor{yellow!25}\textbf{0.937} \\

    \multicolumn{2}{l|}{\textbf{Background Consistency}}
      & 0.871 & \underline{0.881} & \cellcolor{yellow!25}\textbf{0.943}
      & \underline{0.946} & 0.923 & \cellcolor{yellow!25}\textbf{0.949}
      & 0.799 & \textbf{0.918} & \cellcolor{yellow!25}\underline{0.911}
      & 0.901 & \underline{0.935} & \cellcolor{yellow!25}\textbf{0.944}
      & 0.879 & \underline{0.914} & \cellcolor{yellow!25}\textbf{0.937} \\

    \multicolumn{2}{l|}{\textbf{Motion Smoothness}}
      & \underline{0.983} & 0.976 & \cellcolor{yellow!25}\textbf{0.989}
      & \textbf{0.989} & 0.988 & \cellcolor{yellow!25}\underline{0.987}
      & \underline{0.990} & \textbf{0.995} & \cellcolor{yellow!25}0.976
      & 0.954 & \textbf{0.963} & \cellcolor{yellow!25}\underline{0.955}
      & \underline{0.979} & \textbf{0.981} & \cellcolor{yellow!25}0.977 \\

    \multicolumn{2}{l|}{\textbf{Aesthetic Quality}}
      & 0.441 & \underline{0.453} & \cellcolor{yellow!25}\textbf{0.675}
      & \textbf{0.681} & 0.557 & \cellcolor{yellow!25}\underline{0.671}
      & 0.422 & \underline{0.444} & \cellcolor{yellow!25}\textbf{0.659}
      & 0.529 & \underline{0.583} & \cellcolor{yellow!25}\textbf{0.696}
      & \underline{0.518} & 0.509 & \cellcolor{yellow!25}\textbf{0.676} \\

    \multicolumn{2}{l|}{\textbf{Image Quality}}
      & \underline{0.459} & 0.375 & \cellcolor{yellow!25}\textbf{0.782}
      & \underline{0.600} & 0.435 & \cellcolor{yellow!25}\textbf{0.647}
      & \underline{0.378} & 0.215 & \cellcolor{yellow!25}\textbf{0.710}
      & 0.391 & \underline{0.500} & \cellcolor{yellow!25}\textbf{0.695}
      & \underline{0.457} & 0.381 & \cellcolor{yellow!25}\textbf{0.708} \\

    \multicolumn{2}{l|}{\textbf{Mean Score}}
      & \underline{0.710} & 0.697 & \cellcolor{yellow!25}\textbf{0.858}
      & \underline{0.826} & 0.757 & \cellcolor{yellow!25}\textbf{0.843}
      & 0.641 & \underline{0.679} & \cellcolor{yellow!25}\textbf{0.845}
      & 0.727 & \underline{0.770} & \cellcolor{yellow!25}\textbf{0.842}
      & 0.726 & \underline{0.726} & \cellcolor{yellow!25}\textbf{0.847} \\
    \bottomrule
  \end{tabular}}
\end{table}

\begin{table}[tb]
  \caption{\textbf{Trajectory estimation accuracy of rendered videos.} We estimate camera poses from the rendered videos using DA3 and compare them against the reference trajectories, serving as a proxy for the success of spatial scene reconstruction.}
  \label{tab:tra2}
  \centering
  \setlength{\tabcolsep}{3pt}
  \renewcommand{\arraystretch}{1.05}
  \resizebox{\textwidth}{!}{
  \begin{tabular}{cc|ccc|ccc|ccc|ccc|ccc}
  \toprule
\multicolumn{2}{c|}{\textbf{Scene}}
& \multicolumn{3}{c|}{\textbf{Buddha}}
& \multicolumn{3}{c|}{\textbf{Cat}}
& \multicolumn{3}{c|}{\textbf{Corgi}}
& \multicolumn{3}{c|}{\textbf{Frankenstein}}
& \multicolumn{3}{c}{\textbf{Mean}} \\ \midrule

\multicolumn{2}{c|}{\textbf{Metric / Method}}
& D-NeRF & 4DGS & \cellcolor{yellow!25}\textbf{Ours}
& D-NeRF & 4DGS & \cellcolor{yellow!25}\textbf{Ours}
& D-NeRF & 4DGS & \cellcolor{yellow!25}\textbf{Ours}
& D-NeRF & 4DGS & \cellcolor{yellow!25}\textbf{Ours}
& D-NeRF & 4DGS & \cellcolor{yellow!25}\textbf{Ours} \\
\midrule
\textbf{ATE Mean}       & 
& \underline{0.011} & 0.024 & \cellcolor{yellow!25}\textbf{0.004}
& \underline{0.018} & 0.042 & \cellcolor{yellow!25}\textbf{0.014}
& 0.480 & \underline{0.062} & \cellcolor{yellow!25}\textbf{0.054}
& 0.030 & \underline{0.004} & \cellcolor{yellow!25}\textbf{0.003}
& 0.135 & \underline{0.033} & \cellcolor{yellow!25}\textbf{0.019} \\

\textbf{ATE RMSE}       & 
& \underline{0.012} & 0.027 & \cellcolor{yellow!25}\textbf{0.004}
& \underline{0.020} & 0.046 & \cellcolor{yellow!25}\textbf{0.016}
& 0.552 & \underline{0.069} & \cellcolor{yellow!25}\textbf{0.063}
& 0.034 & \underline{0.005} & \cellcolor{yellow!25}\textbf{0.003}
& 0.154 & \underline{0.037} & \cellcolor{yellow!25}\textbf{0.022} \\

\textbf{Rot Mean (Deg)} &
& 6.198 & \underline{5.637} & \cellcolor{yellow!25}\textbf{1.001}
& 62.680 & \textbf{20.620} & \cellcolor{yellow!25}\underline{27.960}
& 173.000 & \underline{9.912} & \cellcolor{yellow!25}\textbf{5.664}
& 23.980 & \underline{1.840} & \cellcolor{yellow!25}\textbf{0.482}
& 66.460 & \underline{9.503} & \cellcolor{yellow!25}\textbf{8.778} \\

\textbf{RPE Trans Mean} &
& 0.010 & \underline{0.009} & \cellcolor{yellow!25}\textbf{0.003}
& \textbf{0.011} & 0.021 & \cellcolor{yellow!25}\underline{0.012}
& 0.164 & \textbf{0.016} & \cellcolor{yellow!25}\underline{0.030}
& 0.024 & \underline{0.003} & \cellcolor{yellow!25}\textbf{0.002}
& 0.052 & \textbf{0.012} & \cellcolor{yellow!25}\textbf{0.012} \\

\textbf{Align Scale}    &
& 2.159 & \underline{1.433} & \cellcolor{yellow!25}\textbf{1.040}
& \textbf{4.536} & 14.110 & \cellcolor{yellow!25}\underline{6.763}
& 5.026 & \underline{4.513} & \cellcolor{yellow!25}\textbf{4.112}
& 4.204 & \underline{1.117} & \cellcolor{yellow!25}\textbf{1.085}
& \underline{3.981} & 5.293 & \cellcolor{yellow!25}\textbf{3.250} \\ \bottomrule
\end{tabular}}
\end{table}

\subsection{Trajectory Estimation Accuracy Oracle-based Pose Evaluation}
\label{sec:eval_pose}
To evaluate camera-motion fidelity as a quality measure of scene reconstruction, we estimate per-frame camera poses from each rendered novel-view video using DA3 and compare them against the reference trajectory (Tab~\ref{tab:tra2}). 
To resolve the global scale and coordinate-frame ambiguity inherent in monocular pose estimation, we first apply Sim3 alignment on camera centers via Umeyama alignment~\cite{umeyama1991least}, and then evaluate the aligned poses.
We report the following metrics:
\begin{itemize}
\item \textbf{ATE (Mean / RMSE)}: absolute translation error of camera centers with respect to the reference trajectory, reflecting global trajectory consistency.

\item \textbf{Rotation error (Mean, deg)}: geodesic angular difference between predicted and reference rotations, measuring orientation accuracy.

\item \textbf{RPE Trans Mean ($\Delta=1$)}: frame-to-frame relative translation error, capturing local temporal stability of camera motion.

\item \textbf{Align Scale}: the Sim3 scale parameter $s$ obtained from Umeyama alignment, indicating the magnitude of global scale correction required.
\end{itemize}

\begin{figure}[tb]
  \centering
  \begin{subfigure}[!t]{0.41\linewidth}
    \centering
    \includegraphics[width=\linewidth]{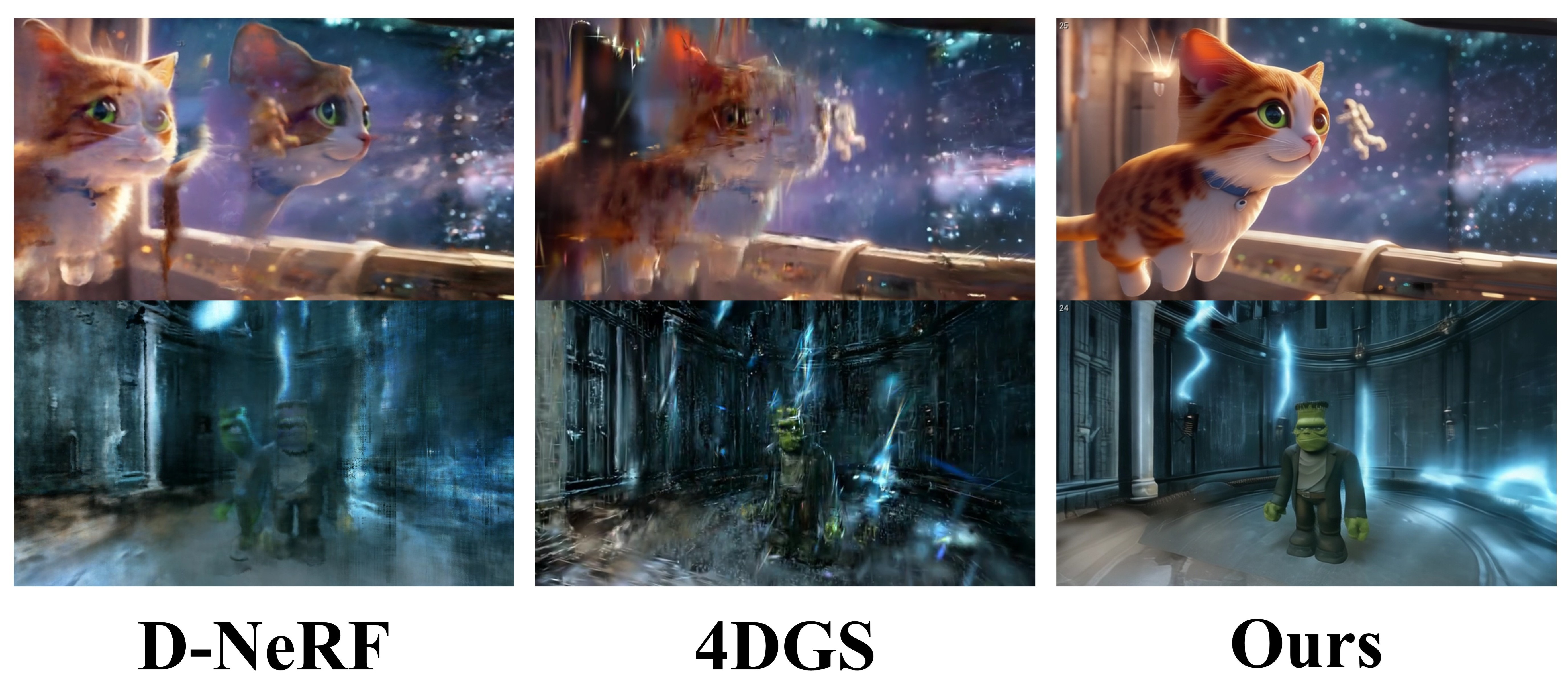}
  \caption{Novel-view rendering from a scene reconstructed from two-view NVS video.}
    \label{fig:self-c0}
  \end{subfigure}\hfill
  \begin{subfigure}[!t]{0.55\linewidth}
    \centering
    \includegraphics[width=\linewidth]{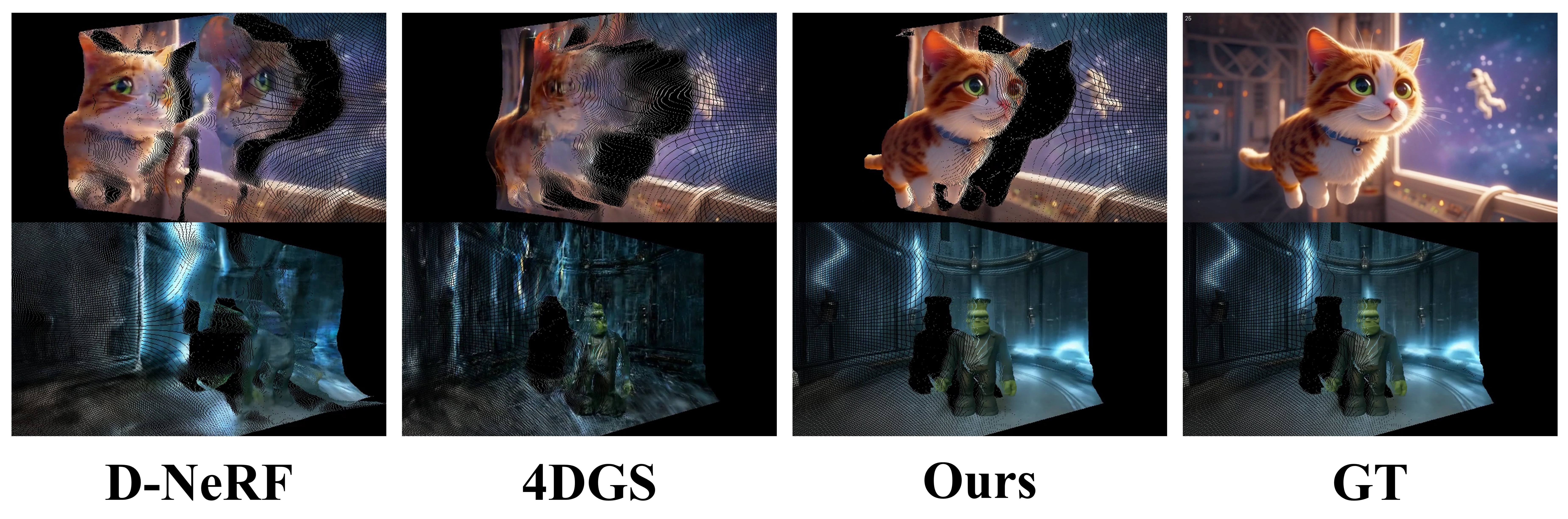}
    \caption{Source-view re-rendering from the novel-view video in (a), used to assess self-consistency.}
    \label{fig:self-c}
  \end{subfigure}
  \caption{\textbf{Qualitative Comparison of Novel-view Rendering of Reconsturcted 4D Scene}}
  \label{fig:qual_two_view_r3econ}
\end{figure}

\subsection{Self-Consistency Evaluation : Using Warped Video}
\label{sec:eval_warp}
We further evaluate geometry--motion consistency via reprojection. Using DA3-estimated depth and pose, each frame of novel view render (Fig.~\ref{fig:self-c0}) is warped to a fixed anchor view and compared against the source-view reference video (Fig.~\ref{fig:self-c}). This evaluation directly tests whether the estimated depth and camera motion can reproject frames into a temporally coherent, view-consistent reconstruction, thereby quantifying geometry--motion consistency.

 As shown in the quantitative evaluations, our method consistently outperforms radiance-field baselines across perceptual quality (Tab~\ref{tab:vbench}), camera-motion fidelity (Tab~\ref{tab:tra2}), and geometric self-consistency (Fig.~\ref{fig:self-c}). In particular, higher VBench scores indicate improved subject and background stability, while lower trajectory errors (ATE/rotation/RPE) demonstrate that the rendered videos better follow the intended camera motion. The self-consistency results further confirm that our geometry refinement produces spatially coherent novel views with reduced cross-view discrepancies.

\begin{figure}
    \centering
    \includegraphics[width=0.95\linewidth]{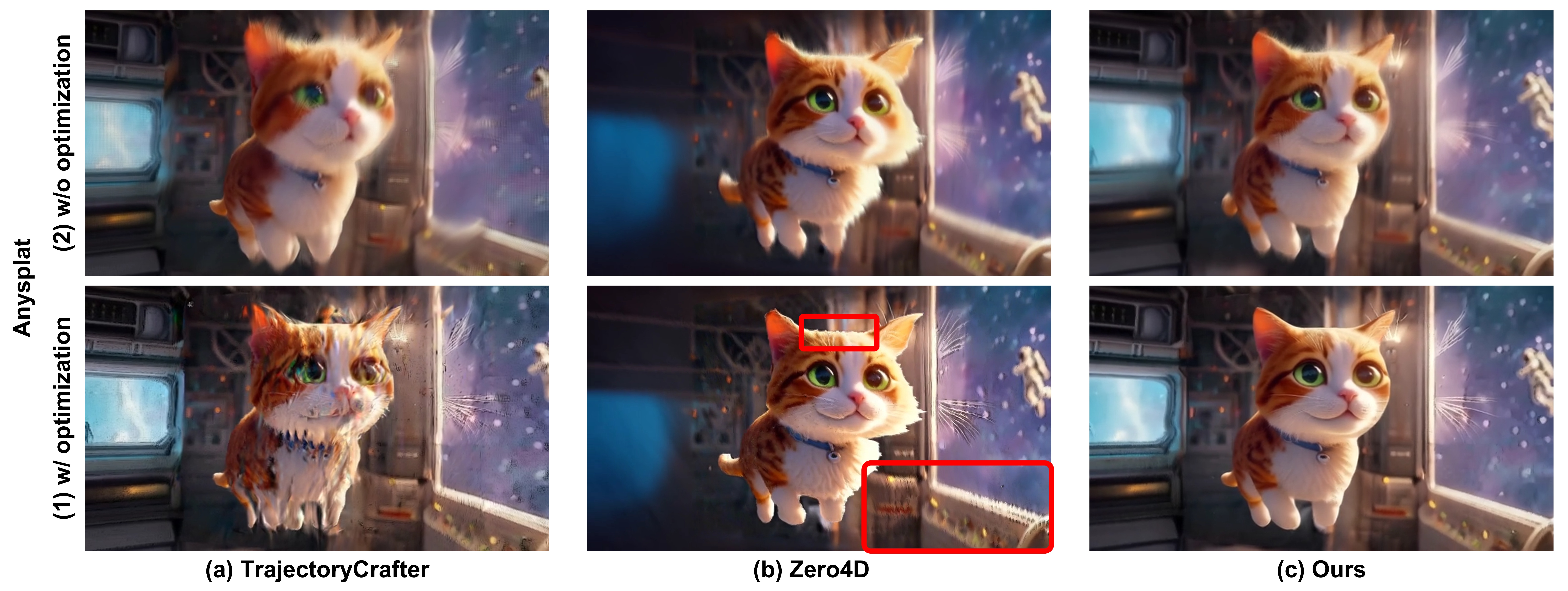}
    \caption{
    \textbf{MVS-induced inconsistency as a bottleneck for 4D reconstruction.}
    We re-render videos from our 4D mesh and reconstruct radiance-based representation using feed-forward Gaussian splatting~\cite{jiang2025anysplat}, with and without per-scene optimization.
    Compared with conventional generation-based pipelines with (a) TrajectoryCrafter~\cite{yu2025trajectorycrafter} and (b) Zero4D~\cite{park2025zero4d}, (c) G4S with 25 views yields more coherent Gaussian Splatting renderings by suppressing view-dependent texture flickering, which appears globally in (a) with 2 views and remains locally in (b) with 25 views during view transition.
    This indicates that generative multi-view inconsistency is a key bottleneck for 4D reconstruction, and that G4S mitigates it by grounding sparse-view generation in consistent 4D geometry.
    }
    \label{fig:fig15}
\end{figure}

\begin{figure}
    \centering
    \includegraphics[width=1\linewidth]{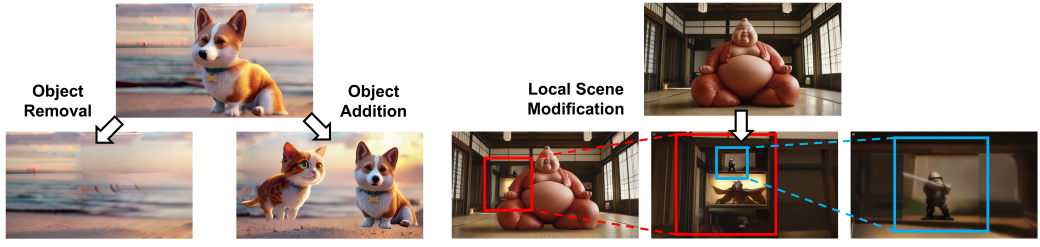}
    \caption{
    \textbf{Downstream Application including Scene Editing.}
    G4S enables stable object removal, addition, and local scene modification.
    }
    \label{fig:scene_edit}
\end{figure}

\section{Conclusion}

We presented \textbf{Generative 4D Stitching}, an explicit framework for constructing coherent 4D scene assets from sparse multi-view video inputs.
Instead of following the conventional generated-data paradigm, which first synthesizes dense multi-view observations and then fits a radiance-based 4D representation through per-scene optimization, our method focuses on resolving the geometric inconsistency introduced by generative novel-view synthesis.

Our key observation is that NVS-generated views should not be used as dense reconstruction supervision, even when they appear visually plausible.
Because their geometry may be misaligned across views, directly lifting them into a 4D asset can propagate camera-depth calibration inconsistency and degrade the final representation.
To address this issue, we identify information-addition regions where new content is genuinely needed, synthesize candidate content only for those regions, and refine the generated geometry by anchoring it to the existing asset geometry.
The refined candidates are then selectively stitched into an explicit 4D representation, so that newly revealed regions can be incorporated without contaminating already reliable geometry.

This formulation avoids relying on under-constrained radiance-based correction under sparse-view settings, while preserving the efficiency of feed-forward geometry lifting.
Experiments show that our approach enables efficient and geometrically consistent 4D scene expansion within the practical capacity of current NVS and feed-forward geometry models.
We hope this work suggests that 4D generation need not always rely on dense generated observations or per-scene radiance-field optimization, and that explicit geometry-aware stitching offers a viable alternative for building coherent 4D assets from sparse observations.

\paragraph{Limitations and Broader Impact.}
G4S builds renderable 4D assets by augmenting unrevealed scene information with generative priors grounded in sparse observations, which may support future research on simulation, AR/VR, scene editing, and multi-observer interaction with dynamic scenes.
However, its performance remains bounded by the NVS backbone and feed-forward geometry estimator, and inaccurate inpainting or geometry may still affect newly added regions.
Moreover, generated unrevealed regions should not be treated as verified real-world reconstructions, since they may contain inaccurate geometry, hallucinated objects, or biased scene details.
Thus, G4S is most suitable for creative, simulation, and research settings rather than safety-critical applications requiring metrically accurate reconstruction.
When applied to real videos, users should consider consent, privacy, and the potential misuse of synthetic 4D media.

{
\small
\bibliographystyle{unsrtnat}
\bibliography{neurips_2026}
}

\newpage
\appendix
\section{Additional Discussion on the Ill-Posed Problem Setting of Generated-Data-Based 4D Generation}

In this section, we provide a more detailed discussion of why generated-data-based 4D generation is fundamentally ill-posed, and where its main bottlenecks arise. A typical generated-data-based 4D generation pipeline can be broadly divided into two stages: (1) \textbf{4D grid generation} and (2) \textbf{4D reconstruction}. In contrast, our target direction is fast 4D reconstruction by leveraging geometry estimated by feed-forward geometry transformers such as Anysplat.

\subsection{Limitations of 4D Grid Generation}
\label{ss.a1}

The first bottleneck lies in the 4D grid generation stage, which aims to construct a spatio-temporal grid of observations. This process is inherently expensive. Since the grid must be populated consistently across both the view and time dimensions, even independently sampling each view-time location already incurs substantial cost. Once an additional refinement process is introduced to enforce geometric and photometric consistency across samples, the overall cost becomes significantly higher.

Meanwhile, independent of this sampling-cost issue, novel-view or multi-view synthesis introduces another limitation. Ideally, generation should only target genuinely unseen regions while remaining consistent with existing observations, which ultimately requires ray-level geometric constraints. However, standard warping-based video NVS pipelines cannot provide such constraints sufficiently. As a result, they may generate visually plausible content that nevertheless conflicts with previous observations. With the inconsistency remaining in the grid, it leads to what we refer to as \emph{inpainting collapse}.

Of course, during the 4D grid generation process, one may partially alleviate this issue by performing additional sampling along trajectories that include the source views, thereby exploiting the VDM prior more extensively. Thus, 4D grid generation itself is not unnecessary. However, despite its partial mitigation effect, it does not fundamentally resolve the problem, while its cost continues to grow. In other words, as one attempts to populate the 4D grid more densely and more consistently, both the computational burden and the structural instability increase together.

\subsection{Limitations of radiance-based 4D Reconstruction}
\label{ss.a2}

\begin{wrapfigure}{r}{0.5\textwidth}
    \vspace{-15pt}
    \centering
    \includegraphics[width=0.5\textwidth]{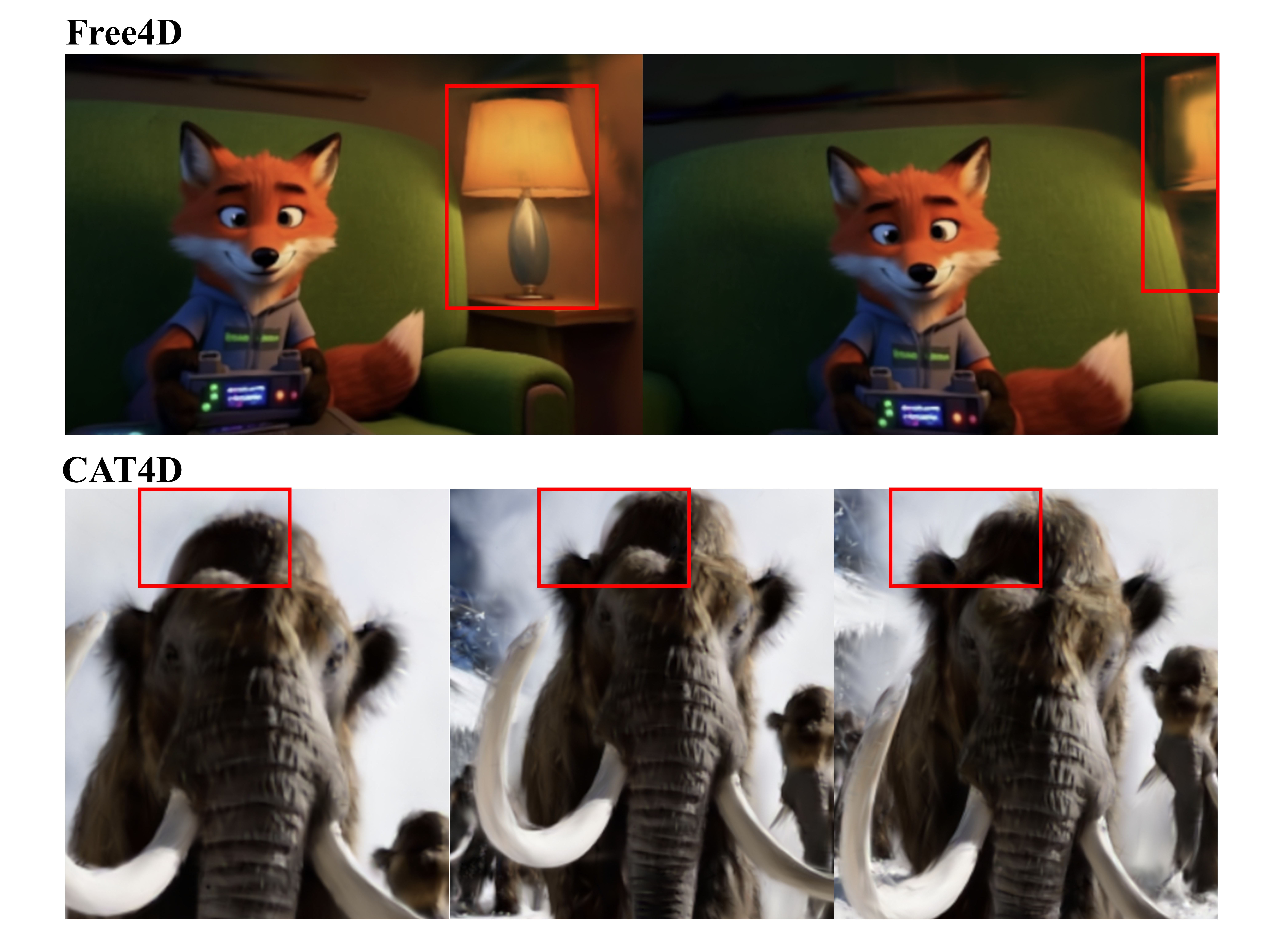}
\caption{
\textbf{Appearance-level absorption of structural mismatch.}
When the underlying scene geometry is not consistently reconciled across views, the residual mismatch tends to manifest through view-dependent appearance changes rather than a stable shared structure by radiance-based optimization. The highlighted regions show this behavior: instead of converging to a coherent geometry, the reconstruction exhibits viewpoint-specific patterns that visually hide, rather than eliminate, the inconsistency.
}
\vspace{-20pt}
\end{wrapfigure}

The next bottleneck arises in the 4D reconstruction stage. Existing approaches typically initialize the scene representation using an estimated geometric prior, such as COLMAP or feed-forward geometry estimators, and then optimize an radiance-based field using multiview photometric loss. In this setting, reconstruction quality strongly depends on two factors: the quality of the initial geometric prior, and the amount and quality of dense views available for photometric supervision.

The problem is that obtaining such dense views requires the preceding 4D grid generation stage itself to become correspondingly denser. That is, improving reconstruction quality demands more views, which in turn requires denser grid sampling, and thus increases the sampling cost accordingly. This cost grows even more rapidly when the goal is not single-view expansion but expansion across multiple views.

Consequently, 4D reconstruction requires dense views for improved performance, while 4D grid generation must become increasingly expensive in order to provide those dense views. In other words, the very condition required for better reconstruction quality also causes the input-generation cost to escalate sharply. This constitutes one of the core sources of ill-posedness in generated-data-based 4D generation.

Moreover, the issue is not merely computational. As shown in the Fig.~\ref{fig:fig9}, even when sufficiently dense views are provided, reconstruction quality can still collapse once the target viewpoint moves beyond the interpolation regime defined by the observed views. Ideally, the optimization process should refine the initial geometry toward a truly consistent solution under multiview photometric supervision. In practice, however, the model often fails to resolve geometric inconsistency itself, and instead relies on view-dependent components such as opacity or spherical harmonics (SH) to dilute the inconsistency. As a result, the model appears to explain all observations, while in fact covering structural disagreement with representational flexibility.

This behavior becomes a clear bottleneck for stable 4D scene modeling, and makes generated-data-based 4D generation fundamentally ill-posed.

\subsection{Current Limits of radiance-based 4D Reconstruction in Sparse-view Settings}

Motivated by the limitations above, we aim to address 4D generation under sparse-view settings rather than relying on dense-view pipelines. To this end, it is important to clarify how existing radiance-based 4D reconstruction methods behave under sparse-view constraints, and why they still tend to converge to sub-optimal solutions.

Radiance-field representations can be broadly divided into NeRF-based and Gaussian-splatting-based formulations, and under sparse-view settings they exhibit different failure modes.

When only two views are available in a generative scene, both NeRF- and Gaussian-splatting-based reconstruction are fundamentally underconstrained, but they tend to fail in different ways. NeRF-based methods, especially without strong geometric priors, may explain the two observations with view-specific or duplicated volumetric structures rather than consolidating them into a single consistent geometry. As a result, the same underlying structure may not be aligned into a unified 3D geometry, but instead appear as fragmented, duplicated, or view-dependent volumetric explanations.

In contrast, Gaussian splatting benefits from explicit geometric initialization, which can reduce severe structural separation by anchoring the reconstruction in 3D space. However, when the two input views are not perfectly geometrically and photometrically consistent, the initialized explicit structure may still be inaccurate. In such cases, the reconstruction may still appear plausible near the observed views, but the inconsistency becomes much more exposed at interpolated or intermediate viewpoints, leading to a rapid degradation in visual fidelity.

In short, under sparse-view constraints, NeRF tends to fail by absorbing ambiguity into diffuse or duplicated volumetric structures, whereas Gaussian splatting tends to fail by explicitly anchoring an imperfect structure whose errors become more directly visible in unseen intermediate views. (Fig. \ref{fig:qual_two_view_r3econ})

\subsection{Information Limits of Sparse-view Observation}

\begin{figure}
    \centering
    \includegraphics[width=1\linewidth]{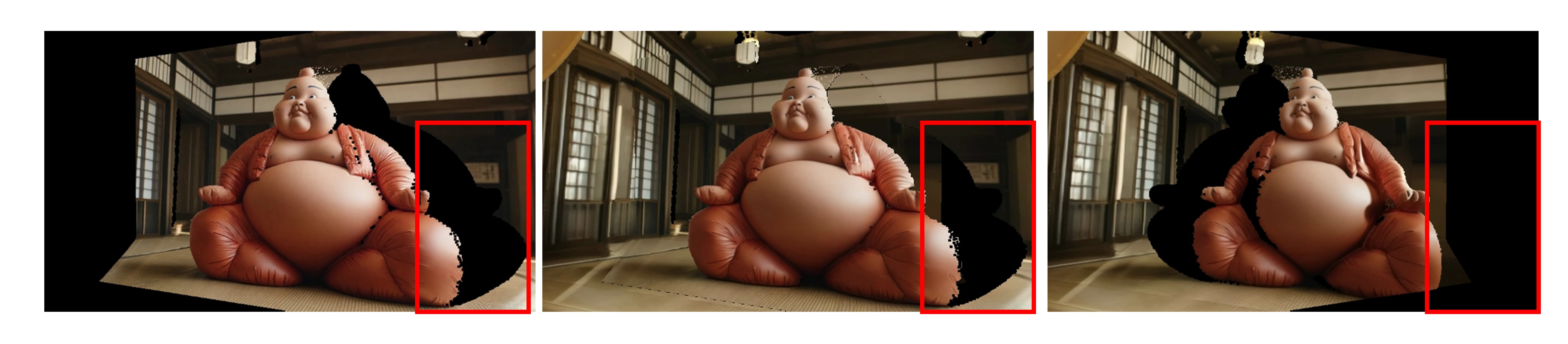}
    \caption{
\textbf{Occlusion-induced missing regions between sparse views.}
Due to foreground occluders, certain background regions remain unobserved across neighboring views. Consequently, interpolated viewpoints between sparse observations contain large information gaps, highlighted in red, which cannot be recovered by simple view interpolation.
}
    \vspace{-10pt}
    \label{fig:fig8}
\end{figure}

Beyond the optimization behavior of reconstruction models, sparse-view 4D generation also suffers from a more fundamental limitation from an information perspective.

When the viewpoint moves, the newly observed camera frustum introduces additional spatial information corresponding to regions that do not overlap with previously observed frusta. In principle, by gradually moving the camera and accumulating such newly revealed regions, the observable portion of the scene expands, allowing the reconstruction system to gather sufficient spatial information over time.

In practice, however, this frustum-level expansion is severely limited by geometric complexity, primarily due to occlusion. If we assume that the background defines the depth limit of a frustum and serves as the natural boundary for view expansion, foreground occluders frequently block portions of the background. As a result, the regions hidden behind these occluders (i.e., occludees) remain unobserved in neighboring views.

Consequently, although the camera frusta from sparse views may partially overlap, the regions that become newly visible between them are often significantly under-constrained. In particular, the interpolated viewpoints between sparse observations tend to contain large information gaps caused by occlusion. In other words, the lack of information in intermediate views is not merely due to insufficient sampling density, but fundamentally arises from occlusion-induced visibility limitations.

To alleviate this issue, we introduce a background completion step that fills the regions behind foreground occluders before performing frustum-level scene expansion. By completing the background geometry behind occluders, the newly generated views become more consistent with the underlying scene structure, enabling more stable frustum-level expansion and reducing the reliance on densely sampled intermediate views.

While this process may still produce secondary occlusions or minor inconsistencies due to imperfect background geometry (e.g., sub-occlusions), these regions are typically small and visually insignificant in most scenes. In practice, they can be sufficiently handled by a lightweight single-step inpainting process without noticeably affecting the overall reconstruction fidelity.

\section{Detailed Implementation of NVS concering Information Addition Area}
\label{app_nvs}

\begin{figure}
    \centering
    \includegraphics[width=1\linewidth]{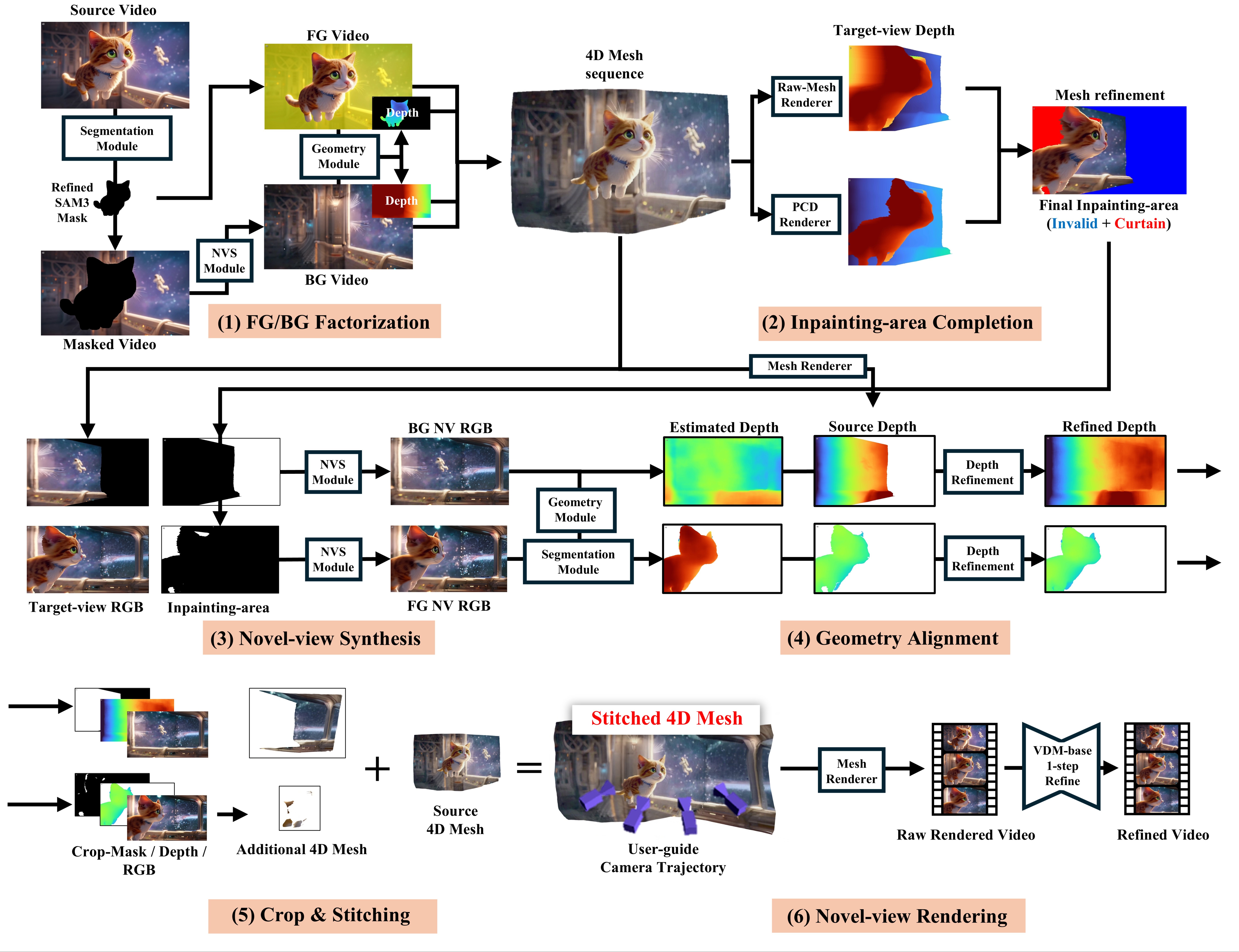}
    \caption{Detailed G4S Pipeline}
    \label{fig:placeholder}
\end{figure}

At time $t$, let the RGB image and depth map of source view $i$ be denoted by
$I_i^t \in \mathbb{R}^{H\times W\times 3}$ and $D_i^t \in \mathbb{R}^{H\times W}$,
respectively, and let the source and target cameras be given by
$\pi_i^t=(K_i^t,R_i^t,\tau_i^t)$ and $\pi_j^t=(K_j^t,R_j^t,\tau_j^t)$.
Here, $K$ denotes the intrinsic matrix, while $R\in SO(3)$ and $\tau\in\mathbb{R}^3$
represent the world-to-camera extrinsics.
Additionally, let $p=(u,v)\in\Omega\subset\mathbb{R}^2$ denote a pixel coordinate in the source image plane,
where $\Omega=\{1,\dots,W\}\times\{1,\dots,H\}$ is the image domain, and let its homogeneous coordinate be
$\bar p=[u,v,1]^\top\in\mathbb{R}^3$.

\paragraph{\textbf{BG/FG Factorization.}}
We first decompose each input frame into foreground and background.
Given an input multi-view video $\{I_i^{t}\}_{i=1}^{N}$ with estimated camera parameters $\{\pi_i^{t}\}$ where $N$ denotes the number of views,
we predict a foreground mask $S_i^{t}\in\{0,1\}^{H\times W}$ using a segmentation model~\cite{carion2025sam3}.
We then obtain a background-only frame by applying a novel-view-synthesis-based inpainting module
$\Phi_{\mathrm{nvs}}$, which fills the masked foreground region using surrounding visual context:
\begin{equation}
B_i^{t} = \Phi_{\mathrm{nvs}}\!\left(I_i^{t},\, S_i^{t}\right),
\end{equation}
where $\Phi_{\mathrm{nvs}}$ denotes the inpainting-based NVS operator such as \cite{yu2025trajectorycrafter}.
This exposes background evidence that would otherwise remain occluded behind the foreground.
As a result, our method can rely less on densely sampled intermediate views while still retaining sufficient background information under sparse-view settings.
It also provides a factorized representation that will later allow us to construct silhouette curtains in a way that is compatible with our BG/FG-factorized NVS pipeline.

\paragraph{\textbf{Silhouette-curtain-based NVS.}}
Most warping-based NVS methods first construct a warped video as a conditioning signal for reprojection by unprojecting source pixels into 3D with an estimated depth map and reprojecting them into the target camera. Based on this warped video, a diffusion prior is then used to complete the target video.

We begin with point-wise unprojection and projection. Given a pixel $p$ with depth $d\in\mathbb{R}_{+}$ under camera $\pi=(K,R,\tau)$,
the corresponding 3D point is recovered by the unprojection operator
$\Pi^{-1}(\pi,p,d)\in\mathbb{R}^3$:
\begin{equation}
X_i^t(p)=\Pi^{-1}\!\left(\pi_i^t,p,D_i^t(p)\right),\quad\text{where }\Pi^{-1}(\pi,p,d)=R^\top\!\left(dK^{-1}\bar p - \tau\right).
\end{equation}
Conversely, point-wise projection   onto the image plane of camera $\pi=(K,R,\tau)$ is defined by
$$q=\mathcal P(\pi,X)$$
where  $X\in\mathbb{R}^3$ denote a 3D point
and
\begin{equation}
q=\left(\frac{\tilde q_x}{\tilde q_z},\frac{\tilde q_y}{\tilde q_z}\right)\in\mathbb{R}^2,
\quad 
\tilde q :=
[\tilde q_x,\tilde q_y,\tilde q_z]= K(RX+\tau),
\end{equation}
where $\tilde q=[\tilde q_x,\tilde q_y,\tilde q_z]^\top\in\mathbb{R}^3$ is the homogeneous image coordinate.

We also denote by
\begin{equation}
z(\pi,X)=\mathbf e_3^\top(RX+\tau), \qquad \mathbf e_3=[0,0,1]^\top,
\end{equation}
the depth of $X$ as seen from camera $\pi$.

These point-wise operators induce two image-level 3D representations of the source frustum.
The first is a point-cloud representation
\begin{equation}
P_i^t
=
\left\{
\left(X_i^t(p),\, I_i^t(p)\right)\;\middle|\; p\in\Omega
\right\},
\end{equation}
which treats each source pixel as an independent 3D sample.
The second is a raw mesh representation $\mathcal M_i^t$ obtained by assigning image-space adjacency to the same unprojected vertices, i.e., by connecting neighboring pixels on the image lattice into triangles.
Thus, while $P_i^t$ is a set of 3D samples, $\mathcal M_i^t$ additionally encodes local surface connectivity over the same lifted geometry.

We now define $\Pi_{i\rightarrow j}^{\mathrm{pcd}}$ and $\Pi_{i\rightarrow j}^{\mathrm{mesh}}$ as the operators that project the source point-cloud and mesh representations into the target view $j$, respectively.
Each operator returns a color rendering, a support mask, and a depth map in the target view:
\begin{equation}
(\tilde I_{i\rightarrow j}^{t,\mathrm{pcd}},\; V_{i\rightarrow j}^{t,\mathrm{pcd}},\; D_{i\rightarrow j}^{t,\mathrm{pcd}})
=
\Pi_{i\rightarrow j}^{\mathrm{pcd}}\!\left(P_i^t;\pi_j^t\right),
\end{equation}
\begin{equation}
(\tilde I_{i\rightarrow j}^{t,\mathrm{mesh}},\; V_{i\rightarrow j}^{t,\mathrm{mesh}},\; D_{i\rightarrow j}^{t,\mathrm{mesh}})
=
\Pi_{i\rightarrow j}^{\mathrm{mesh}}\!\left(\mathcal M_i^t;\pi_j^t\right).
\end{equation}
Here, $V_{i\rightarrow j}^{t,\mathrm{pcd}}$ and $V_{i\rightarrow j}^{t,\mathrm{mesh}}$ indicate which target pixels are supported by the projected point-cloud and mesh representations, respectively,
while $D_{i\rightarrow j}^{t,\mathrm{pcd}}$ and $D_{i\rightarrow j}^{t,\mathrm{mesh}}$ denote the corresponding target-view depth maps.
Point-cloud projection uses bilinear splatting with depth-ordered visibility, whereas mesh projection rasterizes projected triangles and interpolates attributes over covered pixels.
Thus, the former provides point-centered support, while the latter provides surface-centered support.

Most existing warping-based NVS methods directly use the unsupported region
$\bar V_{i\rightarrow j}^{t,\mathrm{pcd}}$ as the inpainting target, yielding
\begin{equation}
\hat I_{i\rightarrow j}^t
=
\Phi_{\mathrm{nvs}}\!\left(\tilde I_{i\rightarrow j}^{t,\mathrm{pcd}},\bar V_{i\rightarrow j}^{t,\mathrm{pcd}}\right),
\qquad
\bar V_{i\rightarrow j}^{t,\mathrm{pcd}} = 1 - V_{i\rightarrow j}^{t,\mathrm{pcd}}.
\end{equation}
While this often produces visually plausible novel views, $\bar V_{i\rightarrow j}^{t,\mathrm{pcd}}$ is merely a visibility-induced hole region and does not faithfully capture the full information-addition region where novel content should be synthesized.
This mismatch is particularly pronounced near foreground occluder silhouettes and coverage gaps.

Our key idea is to detect the information-addition region through the \emph{depth discrepancy} between mesh projection and point-cloud projection.
When the unprojected 3D vertices are connected using raw image-lattice adjacency, the resulting raw mesh may connect points that should not be connected in true 3D geometry.
In particular, near strong depth discontinuities, such raw mesh connections can form thin surface strips spanning foreground and background, naturally inducing silhouette-curtain-like geometry.
By contrast, point-cloud projection only projects independent samples and therefore does not sufficiently cover such thin connecting structures.
As a result, curtain-like regions, which we call {\em silhouette curtain}, appear as discrepancies between mesh and point-cloud projections of the same source frustum (see Fig.~\ref{fig:fig_curtain}) .

We denote the resulting depth-discrepancy-based target-view curtain mask by
\(
C_{i\rightarrow j}^{t,\mathrm{disc}}.
\)

Because our pipeline performs FG/BG-factorized NVS, we additionally introduce a curtain representation that is compatible with this factorized setting. We construct a 3D FG--BG curtain mesh, denoted by $\mathcal C_{i,\mathrm{FB}}^t$, by linking the foreground point and the corresponding background point along the same camera ray for pixels on the boundary ring of the foreground mask.
This FG--BG curtain is not the primary mechanism for curtain detection; rather, it is an auxiliary tool introduced to align curtain construction with our BG/FG-factorized NVS pipeline.
Since $C_{i\rightarrow j}^{t,\mathrm{disc}}$ is already a 2D target-view mask while $\mathcal C_{i,\mathrm{FB}}^t$ is a 3D mesh, the two cannot be merged directly.
We therefore first project the FG--BG curtain mesh into the target view:
\begin{equation}
( \cdot, \;C_{i\rightarrow j}^{t,\mathrm{FB}},\; D_{i\rightarrow j}^{t,\mathrm{FB}})
=
\Pi_{i\rightarrow j}^{\mathrm{mesh}}\!\left(\mathcal C_{i,\mathrm{FB}}^t;\pi_j^t\right),
\end{equation}
where $C_{i\rightarrow j}^{t,\mathrm{FB}}$ denotes the target-view coverage mask of the projected FG--BG curtain and $D_{i\rightarrow j}^{t,\mathrm{FB}}$ its depth map.
The final curtain mask is then defined by
\begin{equation}
C_{i\rightarrow j}^{t}
=
C_{i\rightarrow j}^{t,\mathrm{disc}}
\cup
C_{i\rightarrow j}^{t,\mathrm{FB}}.
\end{equation}

Finally, we define the geometry-guided information-addition mask $M_{i\rightarrow j}^{t}$ and perform NVS inpainting:
\begin{equation}
M_{i\rightarrow j}^{t}
=
\bar V_{i\rightarrow j}^{t,\mathrm{pcd}}
\cup
C_{i\rightarrow j}^{t},
\qquad
\hat I_{i\rightarrow j}^{t}
=
\Phi_{\mathrm{nvs}}\!\left(\tilde I_{i\rightarrow j}^{t,\mathrm{pcd}},\, M_{i\rightarrow j}^{t}\right).
\end{equation}
In this way, synthesis is no longer restricted to visibility-induced holes alone, but is instead aligned with regions where mesh--point-cloud depth discrepancy and FG/BG-factorized curtain geometry both indicate genuine information addition.
The resulting frustum-level NVS outputs are then passed to our geometry stabilization stage (Sec.~\ref{sec:pyramidal_refinement}) and incremental scene stitching stage (Sec.~\ref{sec:geometric_stitching}).

\begin{figure}
    \centering
    \includegraphics[width=1\linewidth]{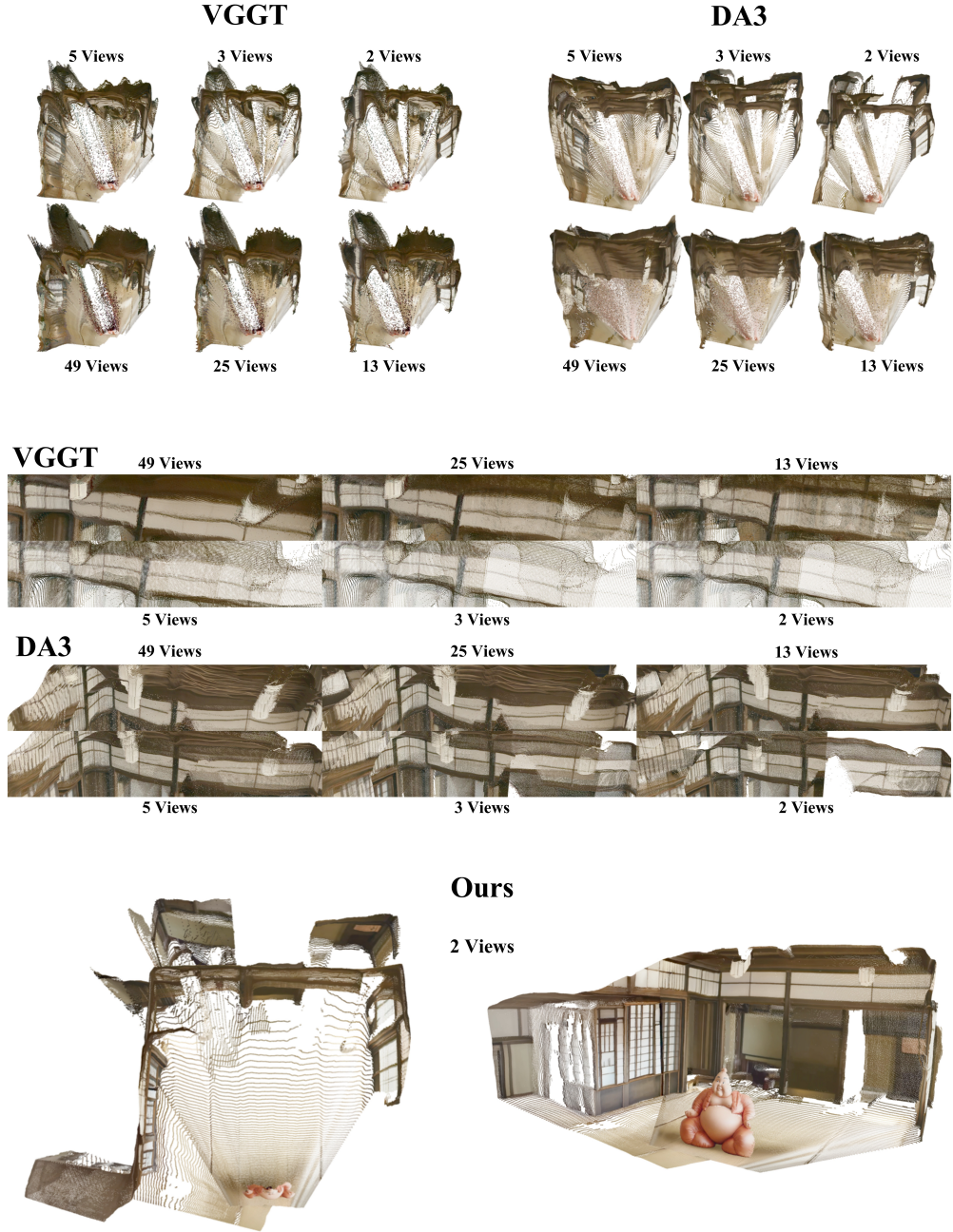}
    \caption{
\textbf{Geometry initialization in generative scenes under varying view counts.}
We compare geometry reconstructed from different numbers of generated views using VGGT and DA3. 
In generative scenes, increasing the number of views does not consistently improve initialization quality, because inpainting collapse introduces cross-view inconsistencies that corrupt multi-view geometric agreement. 
As a result, even denser view sets still exhibit strong structural distortion and unstable surfaces.
In contrast, our method, built on a DA3-based geometry prior, demonstrates substantially higher geometric stability and yields a more coherent reconstruction.
}
    \label{fig:fig9}
\end{figure}

\begin{figure}[!t]
  \centering
    \includegraphics[width=\linewidth]{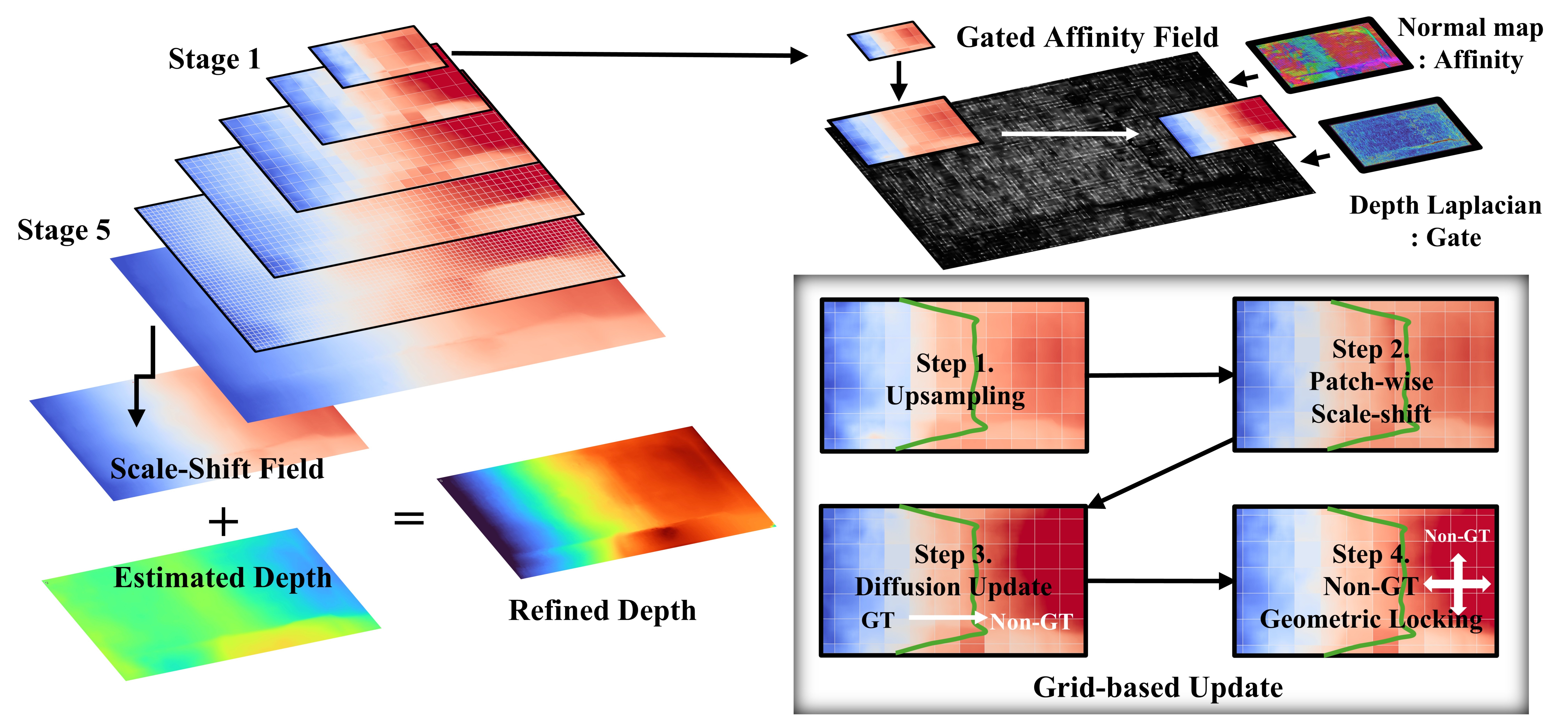}
    \caption{\textbf{Pyramidal Refinement for Source-Anchored Feed-forward Geometry Stabilization.} We construct the continuous scale-shift field constrained to scene geometry for appropriate depth refinement.}
    \label{fig:depth_refine_PYR}
\end{figure}

\begin{figure}[!t]
  \centering
    \includegraphics[width=\linewidth]{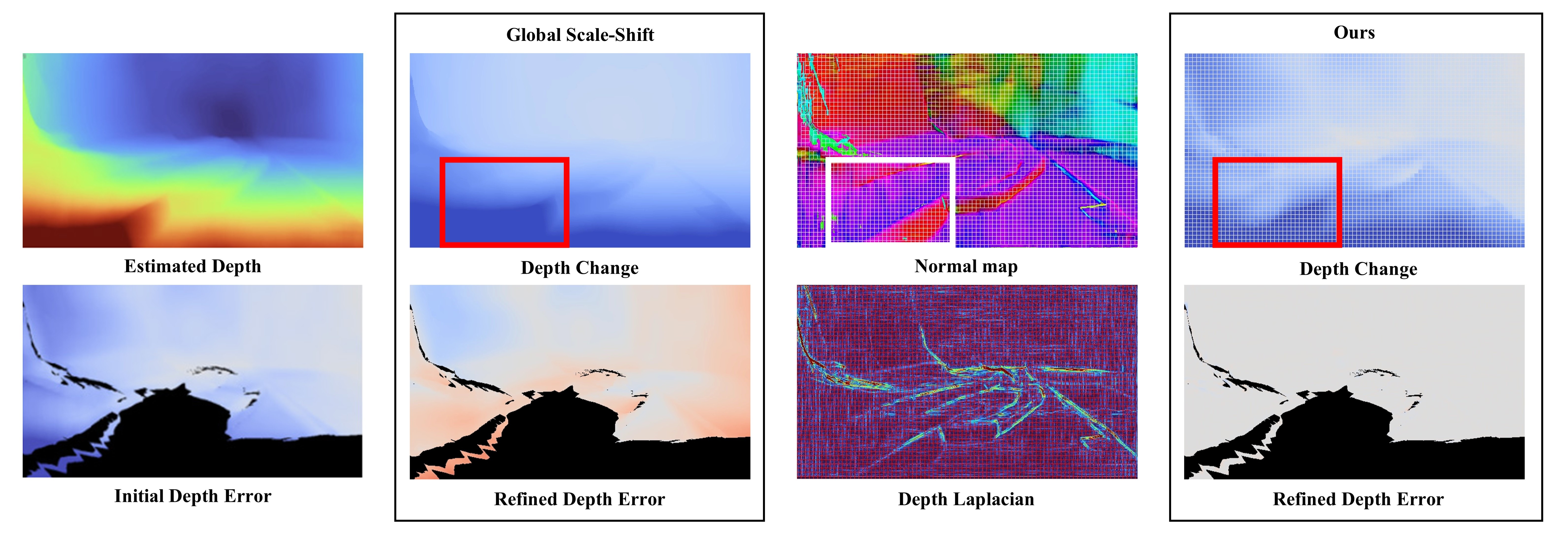}
    \caption{\textbf{Comparison of depth correction behaviors.} The boxed regions show that our method better follows the underlying scene geometry, yielding more consistent refined depth than global scale-shift.}
    \label{fig:depth_refine}
\end{figure}

\section{Detailed Implementation of Depth Refinement}
\label{depth_detail}
\subsection{Implementation Details of Pyramidal Geometry-Aware Local Scale--Shift Refinement}
\label{ss.b1}

This section provides the implementation details of the pyramidal local scale--shift refinement described in Sec.~3.2. While the main text focuses on the motivation and high-level design, here we present the exact formulation of the rendered GT depth, the local anchor parameterization, the stage-wise refinement procedure, the geometry-aware propagation rule, the full-resolution expansion, and the corresponding approximate objective.

We use the TrajectoryCrafter base resolution of $672 \times 384$. Among the FF-GT backbone candidates \{VGGT, PI3, DA3\}, we adopt DA3 in this work. In generative scenes, the NVS poses are determined during sampling, and DA3 is practically convenient because these poses can be explicitly used as conditioning for depth estimation.

The refinement is performed with respect to the rendered GT depth obtained from the source-video constraint. Specifically, the target-view rendered GT depth is obtained by projecting the curtain-excluded source mesh $\mathcal M_{i,\mathrm{gt}}^t$ into the target camera $\pi_j^t$:
\begin{equation}
(\,\cdot,\;V_{i\rightarrow j}^{t,\mathrm{gt}},\;D_{i\rightarrow j}^{t,\mathrm{gt}})
=
\Pi_{i\rightarrow j}^{\mathrm{mesh}}\!\left(\mathcal M_{i,\mathrm{gt}}^t;\pi_j^t\right).
\end{equation}
For brevity, we write
\begin{equation}
D_{\mathrm{gt}} \equiv D_{i\rightarrow j}^{t,\mathrm{gt}},
\qquad
V_{\mathrm{gt}} \equiv V_{i\rightarrow j}^{t,\mathrm{gt}},
\end{equation}
and denote the set of pixels with valid rendered GT depth by
\begin{equation}
\mathcal M_{\mathrm{gt}}=\{\mathbf{x}\mid V_{\mathrm{gt}}(\mathbf{x})=1\}.
\end{equation}

The refinement is performed in a coarse-to-fine pyramidal manner over the following patch grids:
\begin{equation}
128 \rightarrow 64 \rightarrow 32 \rightarrow 16 \rightarrow 8.
\end{equation}
At each stage, we estimate a local scale--shift field on the corresponding patch grid and use it to progressively refine the depth.

At pyramid level $\ell$, let the patch graph be denoted by $\mathcal G^\ell=(\mathcal V^\ell,\mathcal E^\ell)$, and define the patch anchor for node $i$ as
\begin{equation}
\Theta_i^\ell=(s_i^\ell,b_i^\ell),
\end{equation}
where $s_i^\ell$ and $b_i^\ell$ represent the local scale and shift of the corresponding patch. These patch-wise anchors induce a pixel-wise scale--shift field. Specifically, for a pixel $\mathbf{x}$ at level $\ell$, the scale and shift are defined through the anchor-to-pixel blending weights $\alpha_i^\ell(\mathbf{x})$ as
\begin{equation}
s^\ell(\mathbf{x})=\sum_{i\in\mathcal N^\ell(\mathbf{x})}\alpha_i^\ell(\mathbf{x})\,s_i^\ell,
\qquad
b^\ell(\mathbf{x})=\sum_{i\in\mathcal N^\ell(\mathbf{x})}\alpha_i^\ell(\mathbf{x})\,b_i^\ell,
\end{equation}
where $\mathcal N^\ell(\mathbf{x})$ denotes the set of nearby level-$\ell$ anchors for pixel $\mathbf{x}$. The blending weights satisfy
\begin{equation}
\sum_{i\in\mathcal N^\ell(\mathbf{x})}\alpha_i^\ell(\mathbf{x})=1,
\qquad
\alpha_i^\ell(\mathbf{x})\ge 0.
\end{equation}
Using these quantities, the aligned depth at level $\ell$ is written as
\begin{equation}
\tilde D(\mathbf{x};\Theta^\ell)
=
s^\ell(\mathbf{x})\,D^{\mathrm{FF}}(\mathbf{x})+b^\ell(\mathbf{x}).
\end{equation}
Thus, although the optimization variables live at the patch level, the resulting depth correction is realized as a pixel-wise field. Equivalently, the same update can be expressed in residual form as
\begin{equation}
D_{\mathrm{refined}} = D_{\mathrm{base}} + \Delta,
\qquad
\Delta = (s-1)\,D_{\mathrm{base}} + b.
\end{equation}

In practice, instead of solving this field through direct gradient-based optimization, we refine it by repeated stage-wise updates consisting of coarse-to-fine initialization, closed-form anchor fitting, geometry-aware propagation, and invalid-region stabilization.

We first initialize the anchor field at the current level from the previous coarser stage:
\begin{equation}
\Theta_i^\ell \leftarrow \mathrm{Up}\!\left(\Theta^{\ell-1}\right)_i.
\end{equation}
In implementation, we use bilinear interpolation on the patch-center grid. Existing anchor locations retain their previous values, while newly introduced grid points are initialized by interpolation. This step preserves the large-scale correction established at the coarser stage while enabling finer local refinement.

Next, at the current stage, we select patches with sufficient GT support as anchor patches and update their scale and shift through weighted linear fitting. For each patch $P_i^\ell$, let the set of valid GT pixels be
\begin{equation}
\Omega_i^\ell = P_i^\ell \cap \mathcal M_{\mathrm{gt}}.
\end{equation}
Let $x_k = D_{\mathrm{base},k}$, $y_k = D_{\mathrm{gt},k}$, and $w_k$ denote the base depth, rendered GT depth, and fitting weight at each valid GT pixel inside the patch. We define the weighted sums
\begin{equation}
S_w=\sum_k w_k,\qquad
S_x=\sum_k w_k x_k,\qquad
S_y=\sum_k w_k y_k,
\end{equation}
\begin{equation}
S_{xx}=\sum_k w_k x_k^2,\qquad
S_{xy}=\sum_k w_k x_k y_k.
\end{equation}
Then the patch-wise local scale and shift are obtained by the closed-form weighted regression
\begin{equation}
s_i^\ell=\frac{S_w S_{xy}-S_x S_y}{S_w S_{xx}-S_x^2+\epsilon},
\qquad
b_i^\ell=\frac{S_y-s_i^\ell S_x}{S_w+\epsilon},
\end{equation}
where $\epsilon$ is a small constant for numerical stability. The same update can also be interpreted as the solution to the weighted least-squares problem
\begin{equation}
\min_{s_i^\ell,b_i^\ell}
\sum_{\mathbf{x}\in\Omega_i^\ell}
\rho_i^\ell(\mathbf{x})
\left(
s_i^\ell D^{\mathrm{FF}}(\mathbf{x})+b_i^\ell-D_{\mathrm{gt}}(\mathbf{x})
\right)^2.
\end{equation}

To ensure stable anchor estimation, we require the valid GT ratio in each patch (\texttt{min\_unit\_ratio}) to exceed a threshold, which is set to approximately $0.3$ in practice. We further apply MAD-based outlier filtering to reject unreliable anchors, and conservatively restore rejected patches using the average of neighboring valid anchors:
\begin{equation}
\Theta_i^\ell
\leftarrow
\frac{1}{|\mathcal N_i^\ell|}
\sum_{j\in\mathcal N_i^\ell}\Theta_j^\ell.
\end{equation}
Before fitting, we additionally preprocess GT--NVS mismatch regions using line/boundary artifact suppression, depth-disparity-residual-based outlier masking, and removal of small connected components, which reduces the chance that geometrically inconsistent regions contaminate the initial anchor fitting.

After anchor fitting, the estimated scale--shift field is propagated from GT-supported anchor patches to neighboring non-anchor patches. During this step, anchor values remain fixed, and only adjacent non-anchor patches are iteratively updated. The propagation update is written as a weighted relaxation:
\begin{equation}
\Theta_i^\ell
\leftarrow
(1-\eta)\Theta_i^\ell
+
\eta\sum_{j\in\mathcal N_i^\ell}\bar w_{ij}^\ell\,\Theta_j^\ell,
\end{equation}
where $\eta$ is the diffusion mixing coefficient, and the normalized propagation weight is defined as
\begin{equation}
\bar w_{ij}^\ell
=
\frac{w_{ij}^\ell}{\sum_{k\in\mathcal N_i^\ell}w_{ik}^\ell}.
\end{equation}

The propagation should diffuse smoothly within the same local structure, while avoiding indiscriminate spread across structural boundaries. To this end, we define the neighboring-patch weight $w_{ij}^\ell$ as a geometry-aware affinity. We first define the normal similarity as
\begin{equation}
\tilde w^{\mathrm{nrm},\ell}_{ij}
=
\frac{\hat{\mathbf n}_i^{\ell\top}\hat{\mathbf n}_j^\ell+1}{2},
\end{equation}
where $\hat{\mathbf n}_i^\ell$ and $\hat{\mathbf n}_j^\ell$ are unit-normalized depth normals. This term measures whether two neighboring patches share similar local surface orientation.

We also define the Laplacian gate as
\begin{equation}
g^{\mathrm{lap},\ell}_{ij}
=
\mathbf 1\!\left(
\max_{\mathbf{x}\in\mathrm{segment}(\mathbf{c}_i^\ell,\mathbf{c}_j^\ell)}
\left|\nabla^2 D^{\mathrm{FF}}(\mathbf{x})\right|
<
\tau_L
\right),
\end{equation}
which checks whether the path between the two anchor centers crosses a strong depth discontinuity. The hard gate based on normal disagreement is defined as
\begin{equation}
g^{\mathrm{nrm},\ell}_{ij}
=
\mathbf 1\!\left(
\tilde w^{\mathrm{nrm},\ell}_{ij}\ge\tau_n
\right).
\end{equation}

Because applying strict geometric gating from the beginning may overly localize the propagation, we use a two-stage schedule. In the warm-up stage, propagation remains permissive and uses only the soft normal similarity:
\begin{equation}
w_{ij}^\ell
=
\tilde w^{\mathrm{nrm},\ell}_{ij}.
\end{equation}
In the strict stage, both the Laplacian gate and the normal gate are applied so that only geometry-consistent edges are preserved:
\begin{equation}
w_{ij}^\ell
=
g^{\mathrm{lap},\ell}_{ij}\,
g^{\mathrm{nrm},\ell}_{ij}\,
\tilde w^{\mathrm{nrm},\ell}_{ij}.
\end{equation}
In addition, as the refinement proceeds from coarse to fine stages, the Laplacian threshold is progressively tightened so that later propagation becomes increasingly boundary-aware.

After propagation, we perform an additional regularization step on the same set of non-anchor patches. Its purpose is to prevent the refined field from deviating excessively from the relative geometry originally preserved by the FF-GT estimate. In other words, while Step 3 transfers anchor information outward, this step stabilizes invalid or weakly constrained patches. We perform diffusion only within non-GT neighborhoods while keeping GT-aligned regions unchanged:
\begin{equation}
\Theta_i^\ell
\leftarrow
(1-\eta_{\mathrm{ng}})\Theta_i^\ell
+
\eta_{\mathrm{ng}}
\sum_{j\in\mathcal N_i^\ell\,:\,(i,j)\in\mathcal E^\ell_{\mathrm{non\text{-}gt}}}
\bar w_{ij}^\ell\,\Theta_j^\ell.
\end{equation}
Here, $\mathcal E^\ell_{\mathrm{non\text{-}gt}}\subseteq\mathcal E^\ell$ denotes the set of edges whose two endpoints both lie in non-GT regions. The same Laplacian- and normal-based geometric constraints are used in this step as well.

After all pyramid updates are completed, we expand the final anchor field $\Theta^L=\{(s_i^L,b_i^L)\}$ to pixel resolution to obtain the full-resolution residual correction. We again use anchor-to-pixel blending weights $\alpha_i^L(\mathbf{x})$, but at the final level these weights are instantiated as a structure-aware soft assignment. For each pixel $\mathbf{x}$, we define the Laplacian barrier between $\mathbf{x}$ and a nearby final-level anchor center $\mathbf{c}_i$ as
\begin{equation}
\ell_i(\mathbf{x})
=
\max_{\mathbf{y}\in\mathrm{segment}(\mathbf{x},\mathbf{c}_i)}
\left|\nabla^2 D^{\mathrm{FF}}(\mathbf{y})\right|.
\end{equation}
Using this quantity, we construct a soft assignment that assigns larger weights to anchors whose connecting paths cross weaker discontinuities:
\begin{equation}
\alpha_i^L(\mathbf{x})
=
\frac{\exp\!\left(-\beta\,\ell_i(\mathbf{x})\right)}
{\sum_{k\in\mathcal N^L(\mathbf{x})}\exp\!\left(-\beta\,\ell_k(\mathbf{x})\right)},
\qquad
\sum_{i\in\mathcal N^L(\mathbf{x})}\alpha_i^L(\mathbf{x})=1.
\end{equation}
Here, $\mathcal N^L(\mathbf{x})$ denotes the set of nearby final-level anchors for pixel $\mathbf{x}$, and $\beta>0$ controls the sharpness of the soft assignment. For brevity, we may write $\alpha_i(\mathbf{x})\equiv\alpha_i^L(\mathbf{x})$.

The resulting full-resolution residual correction is defined as
\begin{equation}
\Delta^{\mathrm{full}}(\mathbf{x};\Theta^L)
=
\left(
\sum_{i\in\mathcal N^L(\mathbf{x})}\alpha_i^L(\mathbf{x})\,s_i^L
-1
\right)D^{\mathrm{FF}}(\mathbf{x})
+
\sum_{i\in\mathcal N^L(\mathbf{x})}\alpha_i^L(\mathbf{x})\,b_i^L.
\end{equation}
The final refined depth is then written as
\begin{equation}
\hat D(\mathbf{x};\Theta^L)
=
\mathrm{clip}\!\left(
D^{\mathrm{FF}}(\mathbf{x})+\Delta^{\mathrm{full}}(\mathbf{x};\Theta^L)
\right).
\end{equation}
In implementation, this correction is applied on valid pixels, optionally with overlap-fill stabilization.

The above stage-wise refinement can be interpreted as approximately implementing the following objective:
\begin{equation}
\begin{split}
\mathcal L(\Theta)
&=
\lambda_1
\sum_{\mathbf{x}\in\mathcal M_{\mathrm{gt}}}
\left\|
\tilde D(\mathbf{x};\Theta)-D_{\mathrm{gt}}(\mathbf{x})
\right\|_1
+
\lambda_2
\sum_{(i,j)\in\mathcal E}
w_{ij}\left\|\Theta_i-\Theta_j\right\|_1 \\
&\quad+
\lambda_3
\sum_{(i,j)\in\mathcal E_{\mathrm{non\text{-}gt}}}
w_{ij}\left\|\Theta_i-\Theta_j\right\|_1.
\end{split}
\end{equation}
Here, $\Theta$ denotes the anchor field at the final level unless otherwise specified. The first term enforces agreement with the rendered GT depth where valid observations are available, the second term encourages geometry-aware smoothness across neighboring anchors, and the third term provides additional stabilization within non-GT regions.

The average runtime of the full pipeline including this refinement module is about $0.1$ seconds per frame, and the overall runtime scales approximately linearly with the number of frames $F$. This indicates that the proposed sparse-view-based scene-level geometry refinement is practical within our target runtime budget.

\begin{figure}
    \centering
    \includegraphics[width=1\linewidth]{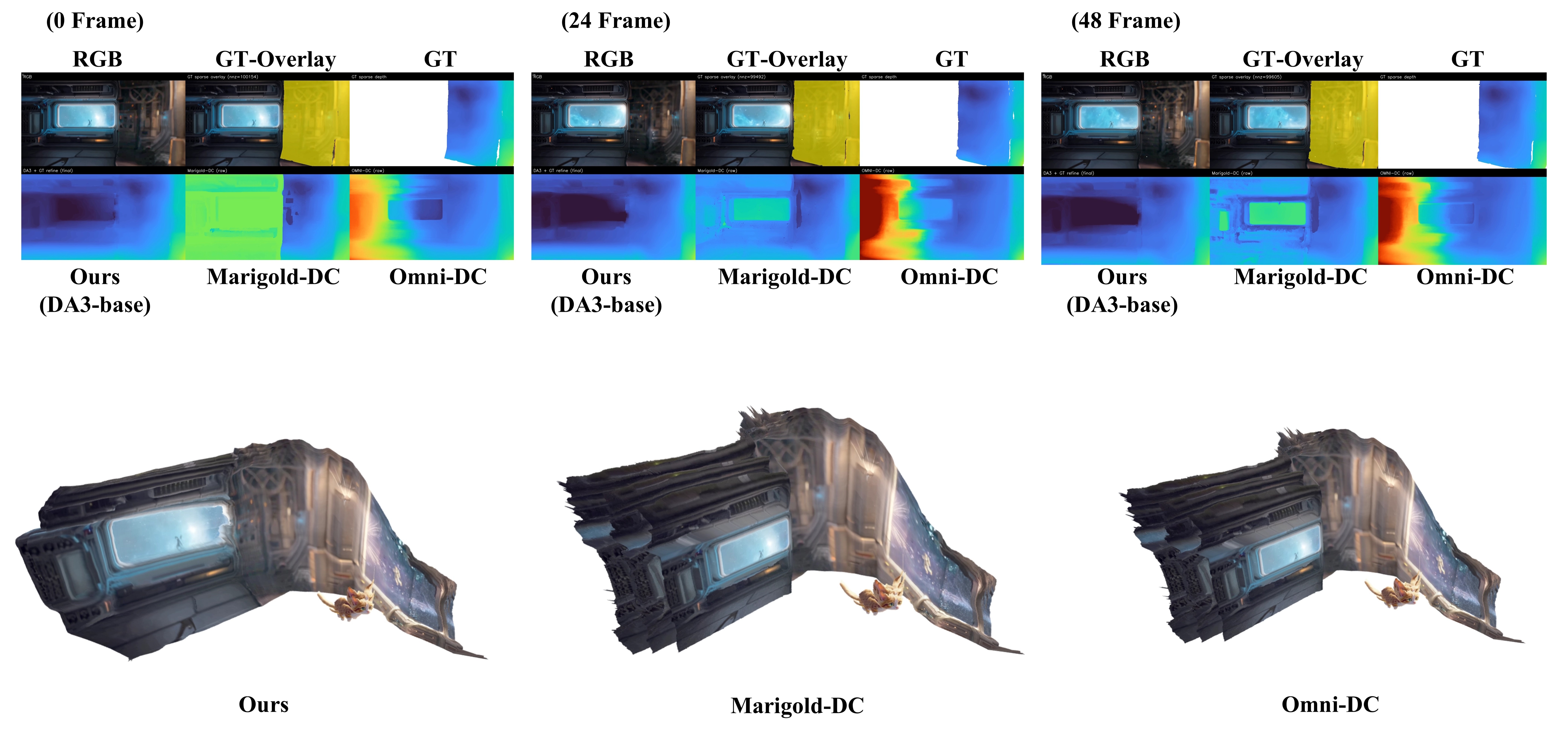}
    \caption{
\textbf{Per-frame accuracy and temporal stability in depth completion.}
We compare our method with image-based depth completion baselines across multiple frames, even when RGB inputs and GT-based geometric cues are provided for each frame. Existing methods suffer from both low per-frame geometric accuracy and strong temporal instability, producing depth predictions that are inaccurate within each frame and inconsistent over time. In contrast, our method yields more accurate per-frame geometry while maintaining stable predictions across frames. The mesh renderings below visualize the accumulated geometric effect of these predictions, showing that our method produces substantially more coherent scene geometry over time.
}
    \label{fig:fig10}
\end{figure}

\subsection{Connection to Depth Completion.}

Our refinement can also be viewed as a form of depth completion, as its goal is to recover missing depth values in regions where ground-truth depth is only partially available. In practice, however, existing depth-completion-based methods fall short of the desired performance in generative scenes: they often fail even at the level of individual frames, and despite being provided with RGB inputs and dense local supervision from partial ground-truth depth, they still do not maintain temporal stability. (Fig. ~\ref{fig:fig10})

\subsection{Additional Pre-processing Implementation Details}
\label{sec:appendix_preprocess_details}

In this section, we describe several auxiliary preprocessing and postprocessing blocks that substantially affect practical stability and output quality.
A common design principle across all of them is to avoid aggressively correcting the entire frame and instead modify only the localized regions where failures are likely to occur.
This allows us to suppress artifacts while preserving the overall structural consistency of the scene. (Fig. ~\ref{fig:fig14})

\begin{figure}
    \centering
    \includegraphics[width=1\linewidth]{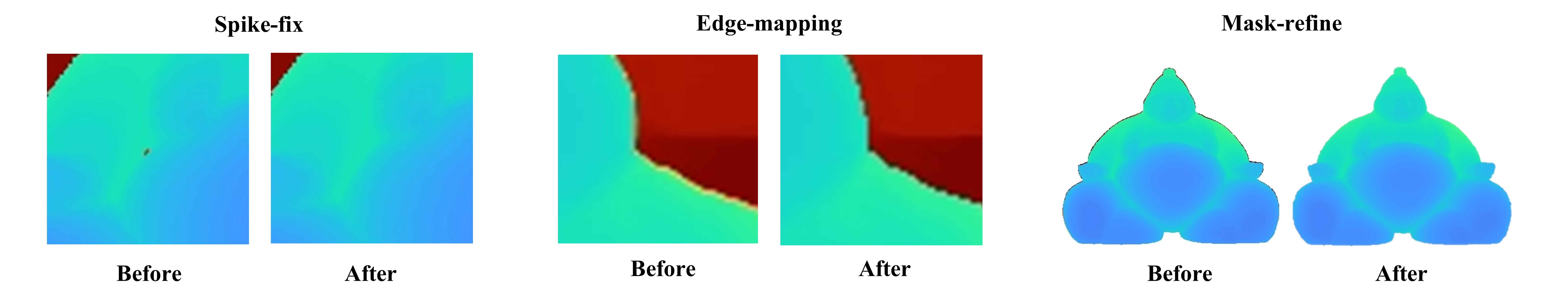}
    \caption{\textbf{Preprocessing refinements for stable geometry estimation.}
Examples of the three preprocessing steps used in our pipeline.
\textbf{Spike-fix}: removes small depth spike outliers using local median-based filtering.
\textbf{Edge-mapping}: reduces anti-aliasing artifacts around depth discontinuities by snapping edge pixels to nearby valid surface samples.
\textbf{Mask-refine}: corrects foreground-background label mixing near occlusion boundaries using local disparity comparison.
Each pair shows the result before and after refinement.}
    \label{fig:fig12}
\end{figure}

\paragraph{Occlusion mask refine.}
Although the occlusion masks predicted by SAM3 are generally useful, foreground and background labels are often mixed near object boundaries.
Since such errors are typically concentrated in a thin band around the mask contour, we do not reclassify the entire mask.
Instead, we only re-evaluate a restricted boundary region.

Specifically, we first construct a thin ring around the mask boundary and define a \emph{cut mask} as the subset of this ring that overlaps with depth edges.
For each pixel inside this cut mask, we compute the mean disparity of the inside and outside regions within a local window, and then reassign the pixel to foreground or background according to which side has the closer disparity mean.
Because this process only updates the ambiguous boundary band, it avoids unnecessarily perturbing the global mask structure.

Finally, we remove small noisy connected components, and preserve the original SAM3 prediction near the image border to prevent over-correction.
In practice, this refinement reduces foreground--background label mixing near occlusion boundaries and improves the stability of the downstream BG/FG separation and geometry lifting stages.

\paragraph{Depth spikefix.}
Feed-forward depth often contains localized spike noise in the form of thin isolated points or small depth islands.
Such outliers can produce floating structures or local geometric corruption during geometry lifting and mesh stitching.
However, treating them with global blur or full-frame smoothing also destroys true boundaries and fine details.
We therefore handle depth spikes as a localized outlier-removal problem.

In practice, we restrict the operation to the mask interior, and use an eroded safe region to avoid unreliable boundary pixels.
For each pixel, we measure its deviation from the local median using a MAD-based criterion, and classify strongly deviating values as outliers.
Only those detected pixels are replaced by the mean of valid neighbors within a local window.
Optionally, this procedure can be limited to small connected components only, so that large structures remain untouched.

This design does not attempt to smooth the entire depth map.
Instead, it only removes the small spike artifacts that are most likely to destabilize later stages, while preserving the overall scene geometry.

\paragraph{Edge mapping.}
When combining DA3 depth with inpainted RGB, anti-aliasing-type artifacts are most pronounced around depth discontinuities.
In particular, object silhouettes and background boundaries often exhibit simultaneous corruption in both RGB and depth, which then propagates instability to refinement and stitching.
To reduce this effect, we explicitly detect depth edges and remap only those pixels.

We first extract depth edges using Laplacian responses, followed by close and dilate operations to stably capture discontinuity regions.
For each edge pixel, we search a small local neighborhood and find a non-edge candidate whose RGB appearance is most similar.
We then copy the RGB and depth values of that candidate to the edge pixel, effectively snapping the problematic boundary pixel to a nearby valid surface.

Because this procedure is applied only at discontinuities, it does not globally alter the frame.
Instead, it locally borrows stable surface values from nearby regions, which reduces tearing and boundary corruption while preserving the global geometric structure.

\paragraph{Curtain depth lower bound.}
Even after depth refinement, some pixels may still be pulled excessively toward the camera.
This often causes background regions to float in front of the foreground or to intrude across boundaries, especially in curtain-based stitched regions.
To suppress this failure mode, we use the curtain-derived depth as a lower bound.

Concretely, whenever the refined depth becomes closer than the curtain depth, we prevent such frontward overshoot by taking the farther of the two depths:
\begin{equation}
D_{\mathrm{out}} = \max\!\left(D_{\mathrm{refined}},\, D_{\mathrm{curtain}}\right).
\end{equation}

In addition, when rendering with the \texttt{nvdiffrast} library, slight staircase artifacts can appear in the rendered curtain depth.
To alleviate this, we optionally smooth the curtain depth with a masked Gaussian filter before applying the lower-bound correction.
We also remove small connected components in the application mask so that the correction does not spread unnecessarily into unrelated regions.

This operation does not introduce new geometry; rather, it acts as a safeguard against physically implausible frontward intrusion.
As a result, it effectively suppresses floating background artifacts and improves the visual consistency of curtain-based geometry augmentation.

\section{Explicit Geometry Representation}

\subsection{Mitigating NVS Limitations through Stitching}
\label{ss.c1}

Our setting reveals a key limitation of NVS that goes beyond imperfect generation in newly exposed regions: it can also break consistency with the projected source evidence itself. In hard generation settings where the source view is not explicitly included in the generation process, the sampled result may drift even in regions already supported by the warped source observation. As a result, generation does not merely complement missing areas, but can also distort structures that should remain consistent with the existing evidence.

This issue is particularly critical in geometry-driven scene expansion, because the warped source image already provides reliable observations over valid projected regions. Ideally, generation should only complement unobserved areas while preserving source-consistent content. In practice, however, hard-conditioned NVS may alter observed and unobserved regions together, weakening geometric consistency with the original view.

To mitigate this issue, we introduce a stitching step that preserves the source-consistent regions from the warped result and uses the sampled image only where genuinely new content is required. This prevents already supported observations from being unnecessarily modified by the generated result, while also relaxing the quality requirement imposed on NVS. In other words, NVS no longer needs to maintain uniformly high fidelity over the entire frame to be useful for scene expansion. By restricting its role to truly novel regions, stitching effectively lowers the minimum quality threshold at which an NVS model can still be practically usable.

\subsection{Visual Limitations of Explicit Geometry Representations}

While avoiding radiance-based optimization allows us to achieve fast and stable reconstruction, explicit geometry representations introduce their own trade-offs. In particular, explicit mesh-based representations tend to produce less natural visual boundaries compared to radiance fields. Since the geometry is discretized into triangles, high-frequency boundary regions may appear jagged or irregular even when the underlying depth estimation is reasonably accurate.

To mitigate this effect, we apply a directional triangle adjustment step based on image-space Sobel filtering, which encourages mesh edges to align with strong image gradients. Although this improves the overall structural alignment between geometry and image boundaries, small irregularities may still remain along object contours due to discretization artifacts.

To further address these residual artifacts, we again leverage a diffusion prior. Specifically, we render the reconstructed mesh and use the resulting images as conditions for a single-step diffusion refinement using our previously employed NVS module, TrajectoryCrafter. This process performs mild image-space regularization, filling micro sub-holes and smoothing high-frequency boundary artifacts while preserving the underlying geometry.

Due to the autoregressive nature of the NVS model, we observe that the generation quality tends to degrade in later frames. To mitigate this issue, we perform bidirectional resampling by providing the frame sequence in both forward and reverse orders. The two generated sequences are then concatenated by taking the first 24 frames from the forward pass and the last 25 frames from the reverse pass, which effectively stabilizes the overall temporal quality and alleviates the degradation in later frames.

\subsection{Application to radiance-based Reconstruction}
\label{ss.c3}

Although our pipeline focuses on explicit geometry representation, the reconstructed scene can also be beneficial for radiance-based reconstruction methods.

radiance-based reconstruction typically relies on dense multi-view observations to provide sufficient photometric supervision for optimization. When only sparse views are available, the reconstruction problem becomes severely underconstrained, often leading to unstable geometry or view-dependent artifacts.

Our stitched explicit geometry representation provides a natural way to alleviate this limitation. Given sparse input views, we first construct an explicit scene representation through our pipeline. By rendering this geometry, intermediate viewpoints can be obtained without requiring additional generative view synthesis.

These rendered views provide additional photometric observations for radiance-based optimization, effectively increasing the view density while preserving geometric consistency. As a result, radiance-based reconstruction can benefit from improved stability and more reliable geometry supervision. (Fig ~\ref{fig:fig15})

\section{Detailed Experimental Setup and Discussion}

\subsection{Novel-view Rendering Setup}

\begin{figure}
    \centering
    \includegraphics[width=0.5\linewidth]{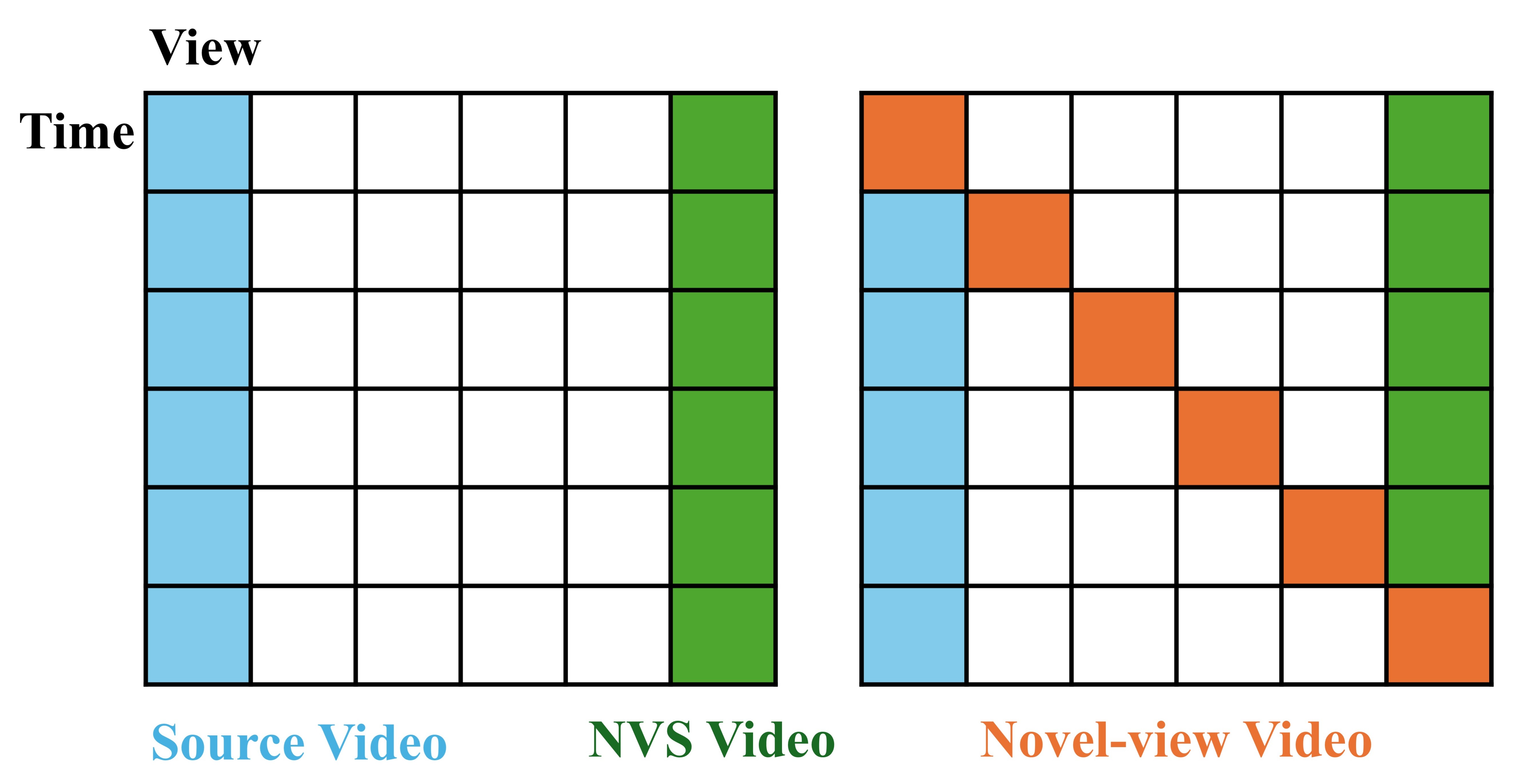}
    \caption{\textbf{Comparison of view–time video layouts.} 
Blue denotes the source video captured from a fixed view over time. 
Green denotes an NVS video rendered at another fixed view for all time steps. 
Orange denotes a novel-view video where the camera view changes over time along a trajectory.}
    \label{fig:nvs}
\end{figure}

For novel-view evaluation, we render intermediate views between two source cameras as illustrated in Fig.~\ref{fig:nvs}. 
Given two source camera poses $(R_0, c_0)$ and $(R_1, c_1)$, we generate interpolated target views by smoothly interpolating both rotation and translation.

For camera rotation, we apply spherical linear interpolation (SLERP) in quaternion space:
\begin{equation}
R(a) = \text{slerp}(q_0, q_1; a),
\end{equation}
where $q_0$ and $q_1$ denote the quaternion representations of $R_0$ and $R_1$, and $a \in [0,1]$ is the interpolation parameter.

For camera position, we use linear interpolation of the camera centers:
\begin{equation}
t(a) = (1-a)c_0 + a c_1.
\end{equation}

The interpolated pose $(R(a), t(a))$ defines the target view used for rendering and evaluation.

\clearpage
\section{Algorithms of Depth refinement}

\begin{algorithm}
\caption{Pyramidal Source-Anchored Depth Refinement on a Patch-Centered Grid}
\label{alg:gs4d-stage}
\KwIn{Feed-forward depth $D^{\mathrm{FF}}\in\mathbb{R}^{F\times H\times W}$, warped GT depth $D_{\mathrm{gt}}$, valid GT mask $M$, pyramid strides $\mathcal{S}=[s_1,\dots,s_K]$}
\KwOut{Refined depth $\hat D$}

Pad $(D^{\mathrm{FF}}, D_{\mathrm{gt}}, M)$ to a multiple of the coarsest stride, with one guard pixel\;
Initialize a patch-centered grid over the image domain\;
Initialize the scale and shift fields $(\mathbf{S}, \mathbf{B}) \gets \varnothing$\;

\For{$k \gets 1$ \KwTo $K$}{
    $s \gets \mathcal{S}[k]$\;
    $(\mathbf{S}, \mathbf{B}) \gets \textsc{Upsampling}(\mathbf{S}, \mathbf{B}, k, s)$\;
    $(\mathbf{S}, \mathbf{B}, \mathbf{A}, \mathbf{F}) \gets \textsc{GT-AnchorFit}(D^{\mathrm{FF}}, D_{\mathrm{gt}}, M, \mathbf{S}, \mathbf{B}, k, s)$\;
    $(\mathbf{S}, \mathbf{B}, \mathbf{A}, \mathbf{\Phi}) \gets \textsc{Prop-to-Non-Anchor}(D^{\mathrm{FF}}, \mathbf{S}, \mathbf{B}, \mathbf{A}, \mathbf{F}, k, s)$\;
    $(\mathbf{S}, \mathbf{B}) \gets \textsc{Non-Anchor-Reg}(D^{\mathrm{FF}}, D_{\mathrm{gt}}, M, \mathbf{S}, \mathbf{B}, \mathbf{\Phi}, k, s)$\;
}

Expand $(\mathbf{S}, \mathbf{B})$ to dense pixel-wise fields $(s_{\mathrm{dense}}, b_{\mathrm{dense}})$ by soft/Laplacian-aware blending\;
$\Delta \gets (s_{\mathrm{dense}} - 1)\odot D^{\mathrm{FF}} + b_{\mathrm{dense}}$\;
$\hat D \gets \Clamp(D^{\mathrm{FF}} + \Delta,\,10^{-6},\,d_{\max})$\;
\Return{$\hat D$}\;
\end{algorithm}

\begin{algorithm}[!t]
\caption{\textsc{Upsampling}: Coarse-to-Fine Grid Initialization}
\label{alg:gs4d-step1}
\KwIn{Previous-level scale and shift fields $(\mathbf{S}^{k-1}, \mathbf{B}^{k-1})$, pyramid level $k$, stride $s_k$}
\KwOut{Current-level fields $(\mathbf{S}^{k}, \mathbf{B}^{k})$}

\eIf{$k = 1$}{
    Keep $(\mathbf{S}^{k}, \mathbf{B}^{k})$ uninitialized; they will be set by GT-supported anchors at the current level\;
}{
    Bilinearly upsample $(\mathbf{S}^{k-1}, \mathbf{B}^{k-1})$ to the current grid resolution\;
}
\Return{$(\mathbf{S}^{k}, \mathbf{B}^{k})$}\;
\end{algorithm}

\begin{algorithm}[!t]
\caption{\textsc{GT-AnchorFit}: Closed-Form Scale-Shift Fitting on GT-Supported Patches}
\label{alg:gs4d-step2}
\KwIn{$D^{\mathrm{FF}}, D_{\mathrm{gt}}, M$, current patch-centered grid $(\mathbf{S}, \mathbf{B})$, pyramid level $k$, stride $s$}
\KwOut{Updated $(\mathbf{S}, \mathbf{B})$, anchor mask $\mathbf{A}$, unresolved GT-supported mask $\mathbf{F}$}

Compute patch radius $r \gets \lfloor s/2 \rfloor$\;
Compute all patch sums by convolution\;

\ForEach{grid cell $g$}{
    Let $x$ denote feed-forward depth values in the patch, $y$ denote warped GT depth values, and $w$ denote valid-pixel weights\;
    $S_w \gets \sum w$\;
    $S_{wx} \gets \sum wx$\;
    $S_{wy} \gets \sum wy$\;
    $S_{wxx} \gets \sum wx^2$\;
    $S_{wxy} \gets \sum wxy$\;
    $\mathrm{denom} \gets S_w S_{wxx} - S_{wx}^2$\;
    $s_g^\star \gets \dfrac{S_w S_{wxy} - S_{wx} S_{wy}}{\mathrm{denom} + \epsilon}$\;
    $b_g^\star \gets \dfrac{S_{wy} - s_g^\star S_{wx}}{S_w + \epsilon}$\;
}

Build a candidate-anchor mask using the valid-ratio test and the denom/finite-value test\;
Apply MAD outlier filtering separately to $\{s_g^\star\}$ and $\{b_g^\star\}$\;
$\mathbf{A} \gets$ candidate $\land$ MAD-inlier(scale) $\land$ MAD-inlier(shift)\;
Assign $(\mathbf{S}, \mathbf{B}) \gets (s_g^\star, b_g^\star)$ on anchor cells in $\mathbf{A}$\;
$\mathbf{F} \gets$ GT-supported grid cells that are not yet anchored\;

\While{there exists a cell in $\mathbf{F}$ that has at least one anchored neighbor}{
    Fill that cell from the average of anchored neighbors in its $3\times 3$ neighborhood\;
    Mark the filled cell as anchored\;
}

\Return{$(\mathbf{S}, \mathbf{B}, \mathbf{A}, \mathbf{F})$}\;
\end{algorithm}

\begin{algorithm}[!t]
\caption{\textsc{Prop-to-Non-Anchor}: Geometry-Aware Relaxation from Anchored to Unanchored Patches}
\label{alg:gs4d-step3}
\KwIn{$D^{\mathrm{FF}}$, scale and shift fields $(\mathbf{S}, \mathbf{B})$, anchor mask $\mathbf{A}$, unresolved GT-supported mask $\mathbf{F}$, pyramid level $k$, stride $s$}
\KwOut{Updated $(\mathbf{S}, \mathbf{B}, \mathbf{A})$, early-propagation flag mask $\mathbf{\Phi}$}

Build neighborhood connectivity and edge weights from Laplacian gating and normal similarity\;
Initialize $\mathbf{\Phi} \gets \mathbf{0}$\;

\For{$i \gets 1$ \KwTo $N_{\mathrm{step3}}^{(k)}$}{
    Select the weight mode: loose warmup or strict geometry gate\;
    Compute the update set $\mathbf{U}$: unanchored cells that have at least one valid anchored neighbor\;

    \If{$\mathbf{U}$ is empty}{
        \KwBreak\;
    }

    \If{$i \le N_{\mathrm{flag}}^{(k)}$}{
        $\mathbf{\Phi} \gets \mathbf{\Phi} \lor \mathbf{U}$\;
    }

    Compute weighted neighbor averages $(\bar{\mathbf{S}}, \bar{\mathbf{B}})$ from anchored neighbors\;
    $\mathbf{S} \gets (1-\alpha_3^{(k)})\mathbf{S} + \alpha_3^{(k)}\bar{\mathbf{S}}$ on $\mathbf{U}$\;
    $\mathbf{B} \gets (1-\alpha_3^{(k)})\mathbf{B} + \alpha_3^{(k)}\bar{\mathbf{B}}$ on $\mathbf{U}$\;
    Promote updated cells to anchors: $\mathbf{A} \gets \mathbf{A} \lor \mathbf{U}$\;
}

\Return{$(\mathbf{S}, \mathbf{B}, \mathbf{A}, \mathbf{\Phi})$}\;
\end{algorithm}

\begin{algorithm}[!t]
\caption{\textsc{Non-Anchor-Reg}: Spatio-Temporal Regularization in Invalid or Weakly Supported Areas}
\label{alg:gs4d-step4}
\KwIn{$D^{\mathrm{FF}}, D_{\mathrm{gt}}, M$, scale and shift fields $(\mathbf{S}, \mathbf{B})$, early-propagation flag mask $\mathbf{\Phi}$, pyramid level $k$}
\KwOut{Regularized $(\mathbf{S}, \mathbf{B})$}

Compute the invalid ratio of each grid cell from patch-valid statistics\;
$\mathbf{R} \gets$ grid cells whose invalid ratio is at least $\tau_{\mathrm{inv}}^{(k)}$ and whose center pixel has valid feed-forward depth\;
Build a 6-neighbor graph on $(t,y,x)$ with 4 spatial neighbors and 2 temporal neighbors\;
Build edge weights:
levels 1--2 use stronger normal-guided weights, whereas finer levels use Laplacian + normal weights\;

\For{$i \gets 1$ \KwTo $N_{\mathrm{step4}}^{(k)}$}{
    Select the weight mode: loose warmup or strict geometry gate\;
    $\mathbf{U} \gets \mathbf{R} \land \text{has-neighbor}$\;

    \If{$i \le N_{\mathrm{freeze}}^{(k)}$}{
        $\mathbf{U} \gets \mathbf{U} \land \neg \mathbf{\Phi}$\;
    }

    \If{$\mathbf{U}$ is empty}{
        \KwContinue\;
    }

    Compute weighted 6-neighbor averages $(\bar{\mathbf{S}}, \bar{\mathbf{B}})$\;
    $\mathbf{S} \gets (1-\alpha_4^{(k)})\mathbf{S} + \alpha_4^{(k)}\bar{\mathbf{S}}$ on $\mathbf{U}$\;
    $\mathbf{B} \gets (1-\alpha_4^{(k)})\mathbf{B} + \alpha_4^{(k)}\bar{\mathbf{B}}$ on $\mathbf{U}$\;
}

\Return{$(\mathbf{S}, \mathbf{B})$}\;
\end{algorithm}


\end{document}